\documentclass{fairmeta}

\usepackage{hyperref}
\usepackage{url}
\usepackage{booktabs}
\usepackage{amsfonts}
\usepackage{microtype}
\microtypesetup{expansion=false}
\usepackage{xcolor}
\usepackage{tikz}
\usepackage{forest}
\usetikzlibrary{trees,positioning}
\usepackage{graphicx}
\usepackage{rotating}
\usepackage{amssymb}
\usepackage{pifont}
\usepackage{multirow}
\usepackage{colortbl}
\definecolor{rowhl}{HTML}{E8F5F2}
\usepackage{comment}
\usepackage{tcolorbox}
\tcbuselibrary{breakable, skins, listings}
\definecolor{codebg}{HTML}{F5F5F5}
\newtcblisting{codeblock}{
  colback=codebg,
  colframe=gray!40,
  boxrule=0.5pt,
  arc=3pt,
  left=6pt, right=6pt, top=4pt, bottom=4pt,
  listing only,
  breakable,
}
\newtcolorbox{promptbox}{
  colback=codebg,
  colframe=gray!40,
  boxrule=0.5pt,
  arc=3pt,
  left=8pt, right=8pt, top=6pt, bottom=6pt,
  fontupper=\small,
  breakable,
}

\newcommand{\cmark}{\ding{51}}
\newcommand{\xmark}{\ding{55}}

\newcommand{\toolscale}{0.9}
\newcommand{\tool}[1]{\scalebox{\toolscale}{\texttt{#1}}}
\newcommand{\SnippetOnly}{\textsc{Snippet-Only}}
\newcommand{\VisitSnippet}{\textsc{Visit-Snippet}}
\newcommand{\VisitSummary}{\textsc{Visit-Summary}}
\newcommand{\OpenFind}{\textsc{Open+Find}}
\newcommand{\FtE}{\textsc{FtE}}

\newcommand{\FtENoGrep}{\FtE{}\,$\setminus$\,\tool{grep}}
\newcommand{\FtENoRead}{\FtE{}\,$\setminus$\,\tool{read}}
\newcommand{\FtESingleFile}{\FtE{}\,$\setminus$\,x-cache}

\definecolor{taxroot}{HTML}{264653}
\definecolor{taxacq}{HTML}{4F86C6}
\definecolor{taxdep}{HTML}{E9A23B}
\definecolor{taxevo}{HTML}{2A9D8F}
\definecolor{taxev}{HTML}{E76F51}
\definecolor{taxsec}{HTML}{8E5EA2}

\title{Fetch-then-Explore: Decoupling Selection from Extraction over a Persistent Workspace for Search Agents}

\author[1]{Qi Liu}
\author[1]{Yiqun Chen}
\author[1]{Zidan Chen}
\author[2]{Yan Gao}
\author[2]{Yi Wu}
\author[2]{Yao Hu}
\author[1]{Jiaxin Mao}
\author[3]{Fengbin Zhu}
\author[3]{Tat-Seng Chua}
\affiliation[1]{Renmin University of China}
\affiliation[2]{Xiaohongshu Inc.}
\affiliation[3]{National University of Singapore}

\abstract{
Search agents now answer questions that take dozens of searches to settle, yet how such an agent reads a page has drawn far less attention than how it finds one. Nearly all of them use one of two document interfaces, and both tie a page to the moment it is opened. \emph{Visit-and-read} injects a reading of the page into the message history at fetch time, fixing that reading before the agent knows which fact it will need. Stateful \emph{browsing} instead extracts on demand from the page in hand, but holds one page at a time and releases it as soon as the agent opens another. Either way, a page that turns out to matter many turns later has to be fetched and rendered into context all over again. We propose \textbf{Fetch-then-Explore}, which separates page selection from evidence extraction and keeps what it selects: pages are recorded in a per-question workspace on the filesystem rather than the context window or a transient session, and evidence is pulled from them on demand later. Selection becomes almost free, extraction can wait until the agent knows what to look for and be repeated as its hypothesis sharpens, and pages are not released when the agent moves on, so evidence accumulates across the trajectory. In a unified ReAct harness with fixed search, we compare Fetch-then-Explore against snippet-only, visit-and-read, and browsing baselines on two open-web benchmarks, BrowseComp and WideSearch, across three agent backbones. It leads BrowseComp accuracy at every backbone and generally matches or exceeds the baselines on WideSearch, and a behavioral analysis traces the gains to the workspace's defining move: returning to a page after leaving it, which it does far more than any transient interface, so evidence missed on a first pass can still be recovered later.
}

\metadata[{\faGithub\ Github}]{\url{https://github.com/liuqi6777/search_agent}}
\correspondence{\email{qiliu6777@gmail.com}, \email{maojiaxin@gmail.com}}

\newif\ifteaser
\teasertrue

\begin{document}

\maketitle

\begin{figure*}[h]
\centering
\includegraphics[width=\textwidth]{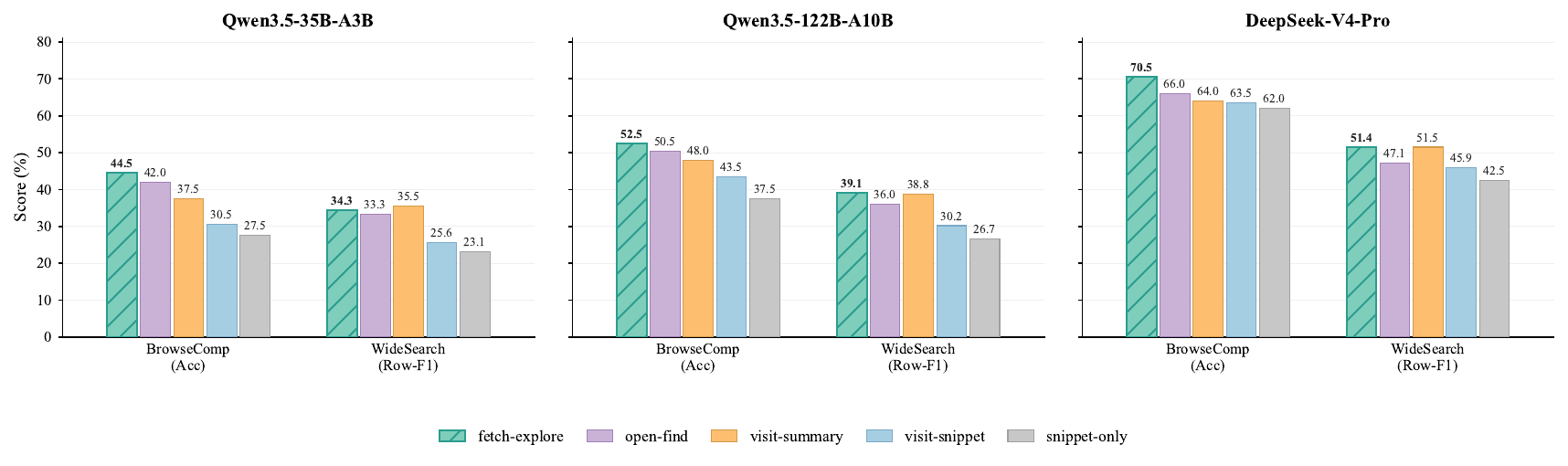}
\caption{Main results across three agents and two open-web benchmarks. \texttt{fetch\_explore}~(ours) is highlighted with a bold outline. Holding search and harness fixed, externalizing document access leads on BrowseComp accuracy at every backbone and stays on par with the strongest visit-and-read baseline on WideSearch Row-F1.}
\label{fig:teaser}
\end{figure*}

\section{Introduction}
\label{sec:introduction}

Long-horizon search agents have become a dominant way to reach the open web for tasks that require gathering evidence across many sources, ranging from open deep-research systems \citep{Li2025WebSailorNavigatingSuper-human, Team2025TongyiDeepResearchTechnical, Qiao2025WebResearcherUnleashingUnbounded, Team2025MiroThinkerPushingthe, Wu2025WebDancerTowardsAutonomous, Li2025WebThinkerEmpoweringLarge, Zheng2025DeepResearcherScalingDeep, Li2025DeepAgentAGeneral, Su2025ScalingAgentsvia, Xi2025ASurveyof} to commercial offerings such as OpenAI Deep Research and Perplexity. Almost all share a ReAct-style architecture \citep{yao2023react}, in which the agent reasons over the message history, issues a tool call, appends the observation, and repeats. Search is the tool such an agent reaches for first and most often, and much of the work on these systems has gone into making that call better.

But search is only the beginning. A result exposes only a title, URL, and snippet, enough to suggest that a page is relevant but rarely enough to answer a multi-hop question or fill a broad table \citep{Wei2025BrowseCompASimple, Wong2025WideSearchBenchmarkingAgentic}. The agent therefore needs a document-access tool that opens a selected result and reveals more of its body. Most open deep-research agents implement a variant of the same interface \citep{Team2025TongyiDeepResearchTechnical, Li2025WebSailorNavigatingSuper-human, Du2026OpenSeekerDemocratizingFrontier}, which we call \emph{visit-and-read}: a \tool{visit}-style call fetches a URL and immediately injects a reading of the page into the message history. Because pages are long and noisy, that reading is not the raw document but a goal-chunked window, an LLM summary, or a truncated body, all chosen \emph{at fetch time}, before the agent knows which fact it will ultimately need. Selecting a page and extracting evidence from it are therefore one action, and its result is fixed on the first pass: whatever falls outside the rendering is out of reach short of fetching the page again. A second family, stateful \emph{browsing} \citep{gpt-oss, Li2026OpenResearcherAFully}, instead \tool{open}s a page into a session first and then \tool{find}s within it, decoupling the two: extraction can wait until the agent knows what to look for and be repeated as the question sharpens. But the session holds one page at a time and releases it on the next \tool{open}, and a \tool{find} reaches only the page in hand, so selected evidence never accumulates.

\begin{figure*}[t]
\centering
\includegraphics[width=\textwidth]{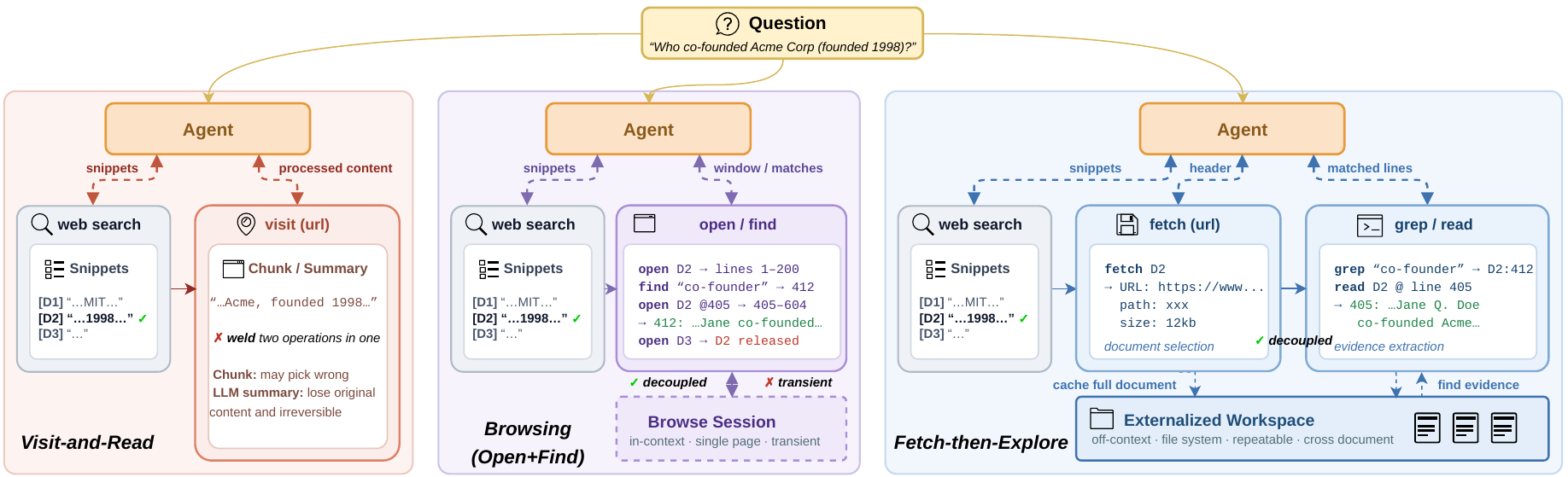}
\caption{\textbf{The document side of a search agent}. \emph{Left, visit-and-read:} \tool{visit} fuses selection and extraction into one fixed fetch-time rendering; everything outside it is gone. \emph{Middle, browsing:} \tool{open}/\tool{find} decouple the two but hold one transient page at a time. \emph{Right, Fetch-then-Explore (ours):} \tool{fetch} records each page in a persistent per-question workspace $\mathcal{C}_q$, deferring extraction to on-demand \tool{grep}/\tool{read} with nothing evicted.}
\label{fig:overview}
\end{figure*}

Over a long-horizon search, an agent may need to look again at a page it has already seen, to verify a claim against it or to read it from a new angle as the question sharpens. Under either interface, the return is expensive: the body is gone once the agent moves on, so coming back means a fresh \tool{visit} or \tool{open}, paying the fetch again and rendering the page into context a second time. To address this challenge, we explore retaining selected pages rather than releasing them, enabling continued use through a persistent storage mechanism.
Such mechanisms already exist in coding agents \citep{yang2024sweagent}, in search systems over closed corpora \citep{Li2026BeyondSemanticSimilarity}, and in recent search agents that externalize state \citep{Jiang2026Harness-1ReinforcementLearning, Zhu2026FS-ResearcherTest-TimeScaling}, but they arrive bundled with other design choices, are applied to a corpus that is already present before the question arrives, or are shipped without measurement.
Whether such a mechanism helps in \emph{open-web} search, where the agent must \emph{find, select, and hold} evidence page by page, remains untested against the interfaces deployed agents actually use.

In light of this, we propose \textbf{Fetch-then-Explore} (Figure~\ref{fig:overview}), which does both: it keeps browsing's decoupling and adds the persistence it lacks, so selected evidence gets a durable home in an external \emph{workspace} that outlives any single page visit. A \tool{fetch} action selects and records a page in a per-question workspace on the filesystem, returning only a lightweight record rather than a one-shot rendering; separate \tool{grep}/\tool{read} actions extract evidence from the stored pages later. Nothing is evicted: the workspace is an explicit, structured artifact that grows over the trajectory and is auditable per question. This yields three properties. First, selection becomes \emph{almost free}: \tool{fetch} puts no page text into context, so a page that merely looks promising can be kept for the cost of a header. Second, extraction becomes \emph{deferred and repeatable}: the agent can query with one hypothesis, read the matching lines, and query again when the hypothesis changes. Third, and unique to the workspace, is \emph{retention}: a page is not released when the agent moves on, so held pages accumulate into a set that stays queryable across documents for the rest of the trajectory.

To test whether these properties translate into end-task gains, we study Fetch-then-Explore in a controlled comparison that varies only the document interface. Because the harness, the search backend, and the backbone are held fixed, any difference traces to how the agent accesses a page rather than how it finds one. We compare it against four baselines spanning the design space of §\ref{sec:method-grid}: a snippet-only floor with no document access, two visit-and-read variants at different fetch-time compression levels, and session-bounded browsing. This lays out two clean contrasts: the visit family tests whether decoupling selection from extraction helps at all, and the already-decoupled browsing baseline isolates the added value of a persistent, cross-document workspace. We run the comparison on two open-web benchmarks, BrowseComp and WideSearch, across three backbones spanning capability tiers. Fetch-then-Explore leads BrowseComp accuracy at every backbone and generally matches or exceeds the baselines on WideSearch, and a behavioral analysis traces its gains to the workspace's defining move: returning to a page after leaving it, which it does far more than any transient interface.

\paragraph{Contributions.} \textbf{(i)} We identify \emph{persistence} as an axis the document interfaces of deployed search agents do not exercise, and instantiate it as \emph{Fetch-then-Explore}: selection decoupled from extraction over an external workspace rather than the context window or a transient session. \textbf{(ii)} We isolate that axis in a controlled comparison against the interfaces deployed agents actually use, holding the harness, the search backend, and the backbone fixed and varying only the document toolset; Fetch-then-Explore leads BrowseComp accuracy at all three backbones and generally matches or exceeds the baselines on WideSearch. \textbf{(iii)} We analyze \emph{why} at the behavioral level, tracing the gains to returning to pages after leaving them and to deferred, repeatable extraction rather than heavier fetching.

\section{Related Work}
\label{sec:related}

\paragraph{Long-horizon search agents.} Open deep-research agents gather evidence across many sources over long horizons, almost all built on the ReAct loop \citep{yao2023react} and extending a longer lineage of retrieval-augmented generation \citep{lewis2020rag}, tool-augmented language models \citep{schick2023toolformer}, classic retrieve-and-reason \citep{press2023selfask, trivedi2023ircot}, browser-assisted question answering \citep{nakano2021webgpt}, and open-web search \citep{Shi2025DeepResearchA}. A fast-growing line trains such agents specifically for web search \citep{Wu2025WebDancerTowardsAutonomous, Li2025WebThinkerEmpoweringLarge, Li2025WebSailorNavigatingSuper-human, Team2025TongyiDeepResearchTechnical, Du2026OpenSeekerDemocratizingFrontier, Li2025DeepAgentAGeneral, Team2025MiroThinkerPushingthe}, pushed by demanding benchmarks: multi-hop BrowseComp \citep{Wei2025BrowseCompASimple} and its variants \citep{Zhou2025BrowseComp-ZHBenchmarkingWeb, Chen2025BrowseComp-PlusAMore}, broad-collection WideSearch \citep{Wong2025WideSearchBenchmarkingAgentic}, and a wave probing contamination and knowledge shortcuts \citep{Fan2026LiveBrowseCompAreSearch, Wang2026EvoBrowseCompBenchmarkingSearch, Deng2025InteractCompEvaluatingSearch}. Across this diversity, the \emph{document} side of the toolset is near-universal in this literature: a visit-style call that \emph{at fetch time} compresses page text into the message history as a goal-ranked window, summary, or truncated body \citep{Team2025TongyiDeepResearchTechnical}. This shared interface, reimplemented by our visit-family baselines (§\ref{sec:method-grid}), has drawn far less scrutiny than the search and training sides, or than the parallel line that folds and rewrites the transcript to keep it inside the context window \citep{Wu2025ReSumUnlockingLong-Horizon, Ye2025AgentFoldLong-HorizonWeb, Chen2025IterResearchRethinkingLong-Horizon}. That line changes what the history keeps, whereas we change whether a page body enters the history at all; we apply no context-management strategy and vary only the document toolset.

\paragraph{Externalized substrates and corpus interaction.} Filesystem-as-working-memory is common in coding agents \citep{yang2024sweagent, wang2024openhands} and in deployed CLI assistants, which expose generic file tools beside their web access and sometimes route tool output through disk \citep{Sen2026IsGrepAll}. The ingredients are therefore already shipped, but the file tools there act on a local project rather than on pages the agent selects from the open web, and no controlled comparison isolates what the substrate contributes. The same principle drives work on agent memory: General Agentic Memory \citep{Yan2025GeneralAgenticMemory} keeps history verbatim in a page-store and compiles context on demand instead of pre-compressing it, while organizing such a store is found to lower retrieval cost without reliably improving answers \citep{Zhou2026Filesystem-BasedMemoryfor}. Both are measured on conversational and embodied memory rather than open-web documents. Search agents have begun to externalize state directly: Harness-1 \citep{Jiang2026Harness-1ReinforcementLearning} hands routine bookkeeping to an environment that maintains a candidate pool and a curated evidence set, and trains the policy inside it with reinforcement learning, while FS-Researcher \citep{Zhu2026FS-ResearcherTest-TimeScaling} archives browsed sources into a file-system knowledge base that a librarian agent builds for a separate report writer. Both find that an external substrate helps, but in both it arrives bundled with training, with an environment-side curation policy, or with a second agent. Fetch-then-Explore instead keeps the decisions in the policy and externalizes only the storage. Stateful browsing \citep{gpt-oss, Li2026OpenResearcherAFully} decouples selection from extraction but holds one transient page at a time; we adopt it as a principal baseline to test whether a persistent, cross-document substrate helps beyond decoupling (§\ref{sec:method-grid}). Direct Corpus Interaction \citep{Li2026BeyondSemanticSimilarity} and follow-ups \citep{Salemi2026GrepSeekTrainingSearch} replace the \emph{retriever} with a filesystem, whereas we replace the \emph{context window}: their filesystem \emph{is} the corpus, static and given in full, so the agent never selects a document and the open web is out of reach. Concurrent shell-over-corpus systems, Dr-DCI \citep{lu2026DrDCI} and RISE \citep{Zhuang2026TowardsRetrievingInteraction}, likewise stay closed-corpus, are evaluated against retrieval baselines rather than the visit-and-read interface deployed agents use, and give no per-stage account of where evidence is lost. The same study finds that the harness and the presentation of tool outputs matter as much as retrieval, reinforcing our choice to hold both fixed and vary only document access.

\paragraph{Select-then-extract in pre-neural QA.} The decomposition we advocate has deep roots: TREC-era retriever--reader pipelines separated passage retrieval from answer extraction \citep{voorhees1999trec8, moldovan2000structure, tellex2003passage, ahn2004wikipedia, chen2017drqa}, and predictive annotation even pushed extraction to index time \citep{prager2000predictive}. But those stages were fixed feed-forward operators run once: type-driven rather than hypothesis-driven, with no retry. The modern visit-and-read interface quietly re-fused them, and browsing re-separates them one transient page at a time; Fetch-then-Explore restores the decomposition over a durable substrate, inside an agentic loop where the policy decides when to extract and can return to held documents as its hypothesis sharpens.

\section{Method}
\label{sec:method}

We first fix a single agent model in which every document interface, or \emph{paradigm}, reduces to a choice of document-access toolset (§\ref{sec:method-agent}). We then present the Fetch-then-Explore (\FtE{}) paradigm, which decouples document selection from evidence extraction and retains fetched content in a persistent per-question cache queried by a small set of on-demand tools (§\ref{sec:method-cache}), and finally situate it and the baselines within a shared space of design dimensions (§\ref{sec:method-grid}).

\subsection{Agent Model}
\label{sec:method-agent}

We instantiate every paradigm inside one ReAct agent \citep{yao2023react} so paradigm-level differences cannot be confounded with prompting or harness variation. Given a question $q$, at each step $t$ the policy $\pi_\theta$ emits a reasoning trace $\tau_t$ and an action $a_t \in \mathcal{A}$ conditioned on the running history $h_t$,
\begin{align*}
(\tau_t, a_t) &\sim \pi_\theta(\cdot \mid h_t), \\
h_t &= \bigl(q,\, \tau_1, a_1, o_1,\, \ldots,\, \tau_{t-1}, a_{t-1}, o_{t-1}\bigr).
\end{align*}
An action is either a tool call or the terminal answer $a_\perp$; a tool call returns an observation $o_t = \mathcal{T}(a_t)$ that is appended to the history, and the rollout ends once the policy emits $a_\perp$ or a step budget is exhausted. The action set splits into three disjoint parts,
\begin{equation*}
\mathcal{A} \;=\; \mathcal{A}_{\mathrm{search}} \;\sqcup\; \mathcal{A}_{\mathrm{doc}} \;\sqcup\; \{a_\perp\},
\end{equation*}
a fixed set of \tool{search} tools $\mathcal{A}_{\mathrm{search}}$, a document-access toolset $\mathcal{A}_{\mathrm{doc}}$, and the terminal answer. A \emph{tool profile} is a choice of $\mathcal{A}_{\mathrm{doc}}$: with $\pi_\theta$, $\mathcal{A}_{\mathrm{search}}$, and the rest of the harness held fixed, every paradigm we study is one such choice (§\ref{sec:method-grid}). Fixing $\mathcal{A}_{\mathrm{search}}$ is deliberate: it bounds what any document-access mechanism could recover, so varying it would conflate \emph{where evidence is held} with \emph{how it is found}.

\subsection{The Fetch-then-Explore Paradigm}
\label{sec:method-cache}

\FtE{} is one such tool profile, built on the two principles motivated in §\ref{sec:introduction}: decouple selection from extraction, and retain what is selected. Where \tool{visit} fuses the two acts and a browsing session separates them one transient page at a time, \FtE{} keeps the separation and adds retention. \emph{Fetch} acquires a document and stores its full body off-context; \emph{explore} queries the stored bodies later, once the hypothesis is sharper.

\paragraph{Tools.} Both principles are realized through one new component: the persistent workspace of §\ref{sec:introduction}, which we implement as a \emph{per-question cache} $\mathcal{C}_q$, a temporary directory holding one plain-text file per fetched resource. Three tools act on it, and their division of labor is the paradigm. The tools themselves are ordinary, the same \tool{fetch}/\tool{grep}/\tool{read} a coding agent has over a repository \citep{yang2024sweagent}; what differs is that a repository is given in full before the task begins, whereas $\mathcal{C}_q$ holds only what this agent chose to keep.

\tool{fetch}(\textit{url}) is the only writer. It retrieves the page, normalizes it to plain text, stores the entire body in $\mathcal{C}_q$ under a filename derived from the URL, and returns only a short record: path, character count, and whether the page is newly fetched or already held. The body never enters the context, so selection costs a fixed handful of tokens however long the page is, and because the filename is deterministic in the URL, fetching a page twice returns the record rather than re-downloading.

\tool{grep}(\textit{pattern}, \textit{path}, \textit{context}) locates evidence with a case-insensitive regular expression. With \textit{path} omitted it searches every file in $\mathcal{C}_q$ at once, the cache-wide scope no session-bound interface can offer: the agent can ask which of its accumulated pages mentions an entity without knowing in advance which one carries it. Up to $20$ matches come back grouped by file as line-numbered hits, each carrying the $\pm$\textit{context} lines around it, three by default and zero for the matching line alone, so a short fact can be read in place.

\tool{read}(\textit{path}, \textit{offset}, \textit{limit}) returns at most $200$ lines of one file starting from a $1$-indexed \textit{offset}, each prefixed with its line number and followed by the offset to continue from. Offsets are lines rather than character positions, so a hit reported by \tool{grep} plugs straight into a follow-up \tool{read} and locate-then-expand becomes a two-call idiom. A housekeeping tool, \tool{list\_fetched}, enumerates the cache. In the notation of §\ref{sec:method-agent}, \FtE{} therefore fixes $\mathcal{A}_{\mathrm{doc}} = \{\tool{fetch}, \tool{grep}, \tool{read}, \tool{list\_fetched}\}$; full schemas are given in Appendix~\ref{app:impl-tools-structured}.

Every observation is thus bounded independently of document length, and retention keeps those bounds cheap rather than lossy: a mis-aimed window is corrected by another query, not another download. The cache is also an \emph{auditable artifact}: its contents are exactly the resources the agent materialized, so every document it selected can be read off after the run. $\mathcal{C}_q$ is per-question; cross-trajectory persistence is orthogonal and not studied here.

\subsection{Design Space and Baselines}
\label{sec:method-grid}

With the harness, search, and backbone held fixed, the profiles differ only in how a document is accessed once it has been selected.
 We organize them along three dimensions. \emph{Substrate} is where a selected body lives, whether the context window, a transient browse session, or the external cache $\mathcal{C}_q$. \emph{Extraction} is whether content is rendered by a fixed operator at fetch time or pulled by bounded on-demand queries. \emph{Query scope} is whether a query reaches only the current page or the whole cache. Table~\ref{tab:profiles} places each profile in this space. The dimensions are not independent: cache-wide scope presupposes an external substrate, and on-demand extraction presupposes a retained body, so the profiles form a lattice rather than a full grid. Two contrasts carry the argument. The \emph{visit-and-read} family, which fuses extraction into selection, tests whether decoupling helps at all, while \emph{browsing} (\OpenFind{}), which already decouples, isolates what a persistent, cache-wide substrate adds beyond it. Where two profiles differ in a single dimension we read the contrast off directly. \OpenFind{} and \FtE{} differ in both substrate and scope at once, so we separate them in two ways: behaviorally, from tool-use statistics, and by ablation, confining \FtE{}'s \tool{grep} to a single held page to fill the missing rung (§\ref{sec:exp-ablation}).

\begin{table}[t]
\centering
\small
\caption{Baselines and the proposed method (highlighted) as coordinates in the three design dimensions defined above: \emph{substrate}, \emph{extraction}, and \emph{query scope}. Tools per profile are listed in the text.}
\label{tab:profiles}
\setlength{\tabcolsep}{3pt}
\begin{tabular}{llll}
\toprule
Profile & Substrate & Extraction & Scope \\
\midrule
\SnippetOnly{}  & ---               & ---              & --- \\
\VisitSnippet{} & context           & fetch-time       & --- \\
\VisitSummary{} & context           & fetch-time       & --- \\
\OpenFind{}     & session           & on-demand        & page \\
\rowcolor{rowhl}\FtE{} & external ($\mathcal{C}_q$) & on-demand & \textbf{cache} \\
\bottomrule
\end{tabular}
\end{table}

\paragraph{Baselines.} Reading down Table~\ref{tab:profiles}, each baseline fixes a different cell of this space:
\begin{itemize}
  \item \SnippetOnly{} fixes $\mathcal{A}_{\mathrm{doc}} = \varnothing$: it adds no document-access tool and serves as the snippet-sufficiency floor, upper-bounding the questions answerable from search snippets alone.
  \item \VisitSnippet{} and \VisitSummary{}, the \emph{visit-and-read} family used by most deployed deep-research agents \citep{Team2025TongyiDeepResearchTechnical, Du2026OpenSeekerDemocratizingFrontier, Li2025WebSailorNavigatingSuper-human}, share $\mathcal{A}_{\mathrm{doc}} = \{\tool{visit}\}$, where a call on a selected document $d$ renders it at fetch time and returns $R(d)$ \emph{into context}. The two variants change only the operator $R$, a goal-conditioned chunk versus an evidence summary from a dedicated \emph{summary} model, and so together isolate the \emph{extraction} operator $R$ at a fixed in-context substrate.
  \item \OpenFind{}, in the style of browser tools \citep{gpt-oss, Li2026OpenResearcherAFully}, takes $\mathcal{A}_{\mathrm{doc}} = \{\tool{open}, \tool{find}\}$ and decouples selection from extraction, but keeps the substrate internal and transient: \tool{open} loads a page into a session and renders a bounded line-numbered window rather than dumping the body, and \tool{find} greps that page and returns matches with surrounding context, all released on the next \tool{open}. Its extraction is thus bounded and on-demand exactly as in \FtE{}, so the two differ only in \emph{substrate and scope}: a transient single page versus a persistent cache queried across pages.
\end{itemize}

\section{Experiments}
\label{sec:experiments}

We organize the study around two questions. 
\begin{itemize}
  \item \textbf{(1)}~Does a persistent workspace improve end-task accuracy over visit-and-read and over browsing, across backbones and benchmarks (§\ref{sec:exp-main})? 
  \item \textbf{(2)}~\emph{How} does it change the agent's document behavior, in particular its ability to return to a page after leaving it and its use of the on-demand extraction tools (§\ref{sec:exp-analysis})?
\end{itemize}

\subsection{Setup}
\label{sec:exp-setup}

\paragraph{Benchmarks.} We evaluate primarily on two open-web deep-research benchmarks, \textbf{BrowseComp} \citep{Wei2025BrowseCompASimple} and \textbf{WideSearch} \citep{Wong2025WideSearchBenchmarkingAgentic}, chosen to stress complementary axes of a search agent. BrowseComp probes \emph{depth}: each of its $1{,}266$ multi-hop questions turns on a single hard-to-find fact that typically takes dozens of searches and page reads to pin down; due to resource constraints, we randomly sample $200$ of them, held fixed across every profile and backbone. WideSearch probes \emph{breadth}: each of its $200$ table-filling tasks requires collecting many fields at high recall from a wide set of pages. A third benchmark, \textbf{BrowseComp-Plus} \citep{Chen2025BrowseComp-PlusAMore}, adds document-level relevance judgments over a fixed corpus (Appendix~\ref{app:triage}).

\paragraph{Models.} To guard against single-model artifacts, we run every profile on three backbones spanning capability tiers: the mixture-of-experts \textbf{Qwen3.5-35B-A3B} and \textbf{Qwen3.5-122B-A10B}~\citep{qwen3.5}, and the stronger \textbf{DeepSeek-V4-Pro}~\citep{deepseekai2026deepseekv4}.

\paragraph{Tools.} Every profile runs in the shared ReAct harness of §\ref{sec:method-agent} and differs only in its document-access tool profile (§\ref{sec:method-grid}); the \tool{search} side is identical throughout. Concretely, \tool{search} is backed by the Serper API\footnote{\url{https://serper.dev/}}, which returns ranked result snippets, and \tool{visit}/\tool{fetch} obtain page content through crawl4ai\footnote{\url{https://github.com/unclecode/crawl4ai}}, which renders HTML and PDF pages to clean text. The \VisitSummary{} profile's fetch-time summarizer is Qwen3.5-35B-A3B~\citep{qwen3.5}, held fixed regardless of the agent backbone so that summary quality is not confounded with backbone strength. All runs share the same per-question budget of up to $300$ turns, and we apply no context-management strategy, so the full interaction history is retained throughout.

\begin{table*}[t]
\centering
\small
\caption{Main results on the two open-web benchmarks across three backbones\ifteaser; the same data is plotted in Figure~\ref{fig:teaser}\fi. Per profile: end-task metrics, mean \tool{search} calls (\emph{\#\textit{Search}}), and document access split into selection (\emph{\#\textit{Select}}) and on-demand extraction (\emph{\#\textit{Extract}}). ``---'': act not performed. \textbf{Bold}/\underline{underline}: best/second within a backbone for the \emph{Acc} and \emph{F1} columns.}
\label{tab:main}
\setlength{\tabcolsep}{3pt}
\begin{tabular}{ll cccc ccccc}
\toprule
& & \multicolumn{4}{c}{\textbf{BrowseComp}} & \multicolumn{5}{c}{\textbf{WideSearch}} \\
\cmidrule(lr){3-6} \cmidrule(lr){7-11}
Backbone & Profile & Acc & \#\textit{Search} & \#\textit{Select} & \#\textit{Extract} & Row-F1 & Item-F1 & \#\textit{Search} & \#\textit{Select} & \#\textit{Extract} \\
\midrule
\multirow{5}{*}{Qwen3.5-35B-A3B}
  & \SnippetOnly{}         & 27.5             & 70.2  & ---  & ---  & 23.1            & 52.3            & 38.9  & ---  & ---  \\
  & \VisitSnippet{}        & 30.5             & 60.3  & 7.4  & ---  & 25.6            & 55.6            & 29.0  & 15.1 & ---  \\
  & \VisitSummary{}        & 37.5             & 58.7  & 9.2  & ---  & \textbf{35.5}   & \underline{63.2}& 29.3  & 17.1 & ---  \\
  & \OpenFind{}            & \underline{42.0} & 58.1  & 5.0  & 1.0  & 33.3            & 63.0            & 24.2  & 8.5  & 1.1  \\
  \rowcolor{rowhl}& \FtE{} (ours) & \textbf{44.5} & 58.3  & 5.4  & 6.7  & \underline{34.3}& \textbf{63.9}   & 21.6  & 11.2 & 12.0 \\
\midrule
\multirow{5}{*}{Qwen3.5-122B-A10B}
  & \SnippetOnly{}         & 37.5             & 76.4  & ---  & ---  & 26.7            & 57.0            & 44.3  & ---  & ---  \\
  & \VisitSnippet{}        & 43.5             & 68.4  & 8.1  & ---  & 30.2            & 59.9            & 29.3  & 13.7 & ---  \\
  & \VisitSummary{}        & 48.0             & 74.5  & 8.8  & ---  & \underline{38.8}& \underline{65.1}& 31.7  & 15.3 & ---  \\
  & \OpenFind{}            & \underline{50.5} & 60.3  & 6.2  & 0.8  & 36.0            & 63.9            & 22.8  & 8.3  & 1.3  \\
  \rowcolor{rowhl}& \FtE{} (ours) & \textbf{52.5} & 62.8  & 7.3  & 7.8  & \textbf{39.1}   & \textbf{65.2}   & 20.3  & 11.3 & 12.4 \\
\midrule
\multirow{5}{*}{DeepSeek-V4-Pro}
  & \SnippetOnly{}         & 62.0             & 101.1 & ---  & ---  & 42.5            & 67.3            & 88.3  & ---  & ---  \\
  & \VisitSnippet{}        & 63.5             & 69.9  & 11.1 & ---  & 45.9            & 71.5            & 62.2  & 20.0 & ---  \\
  & \VisitSummary{}        & 64.0             & 80.8  & 13.1 & ---  & \textbf{51.5}   & \underline{71.9}& 50.4  & 20.1 & ---  \\
  & \OpenFind{}            & \underline{66.0} & 61.2  & 9.5  & 1.9  & 47.1            & 69.0            & 35.2  & 12.1 & 2.2  \\
  \rowcolor{rowhl}& \FtE{} (ours) & \textbf{70.5} & 66.9  & 10.7 & 11.5 & \underline{51.4}& \textbf{72.2}   & 36.0  & 15.3 & 21.0 \\
\bottomrule
\end{tabular}
\end{table*}

\paragraph{Evaluation.} We follow each benchmark's official evaluation protocol and prompts, scoring answers with an LLM judge. On BrowseComp we report end-task accuracy; on WideSearch, row- and item-level F1 of the completed table against the gold table. Each cell is run three times, and we report the best of the three runs; the behavioral statistics of §\ref{sec:exp-analysis} are computed on that same reported run. We use DeepSeek-V4-Pro as the judge model, applied identically to every profile and backbone (Appendix~\ref{app:impl-eval}).

\subsection{Main Results}
\label{sec:exp-main}

Table~\ref{tab:main} reports all five profiles on the two open-web benchmarks across the three backbones\ifteaser, and Figure~\ref{fig:teaser} visualizes the same data\fi. The meaningful comparison is \emph{intra-backbone}, setting \FtE{} against the four baselines at a fixed backbone, since each row's absolute level is bounded by that backbone's own capability. The pattern is consistent: under the same harness, search backend, and backbone, \FtE{} is the best profile in four of the six cells and within a fraction of a point of the best in a fifth, and it leads BrowseComp accuracy at all three backbones. One feature of the table aids reading: document access is split into its two acts. \emph{\#\textit{Select}} counts the materialization call, whether \tool{visit}, \tool{fetch}, or \tool{open}, and \emph{\#\textit{Extract}} the on-demand calls, \tool{grep} and \tool{read} for \FtE{} and \tool{find} for \OpenFind{}. The visit family folds the second act into the first.

\paragraph{The snippet floor sets the headroom.} \SnippetOnly{} answers from the shared \tool{search} stream with no document access, upper-bounding the share of questions on which document access is in principle unnecessary. This floor varies sharply. On BrowseComp it climbs with capability from $27.5$ to $37.5$ to $62.0$; even at DeepSeek-V4-Pro, where the snippet stream alone answers $62.0\%$, document access still adds a clear $8.5$\,pp on top, reaching $70.5$ with \FtE{}. On WideSearch it stays low throughout, at $23$--$43$ Row-F1, because broad-collection tables are essentially unsolvable from snippets.

\paragraph{\FtE{} matches or exceeds visit-and-read.} The clearest result is on BrowseComp, where \FtE{} is the most accurate profile at every backbone, at $44.5$, $52.5$, and $70.5$, ahead of the raw \VisitSnippet{} by $7$ to $14$\,pp and of the second-best profile by $2$ to $4.5$, with the largest margin on the strongest backbone. The ordering there is fixed across all three: \FtE{} $>$ \OpenFind{} $>$ \VisitSummary{} $>$ \VisitSnippet{} $>$ \SnippetOnly{}. On WideSearch the primary Row-F1 is closer: \FtE{} leads at Qwen3.5-122B-A10B and comes within $0.1$ at DeepSeek-V4-Pro, and the one cell where a baseline leads by more than a rounding margin is the weakest backbone, where the strengthened \VisitSummary{} edges it by $1.2$ Row-F1. Against the raw \VisitSnippet{}, \FtE{} leads WideSearch Row-F1 at all three.

The two benchmarks reward different things, which is why the separation is sharper on BrowseComp. WideSearch fills broad tables and so rewards pulling many fields at high recall, which a faithful high-coverage extractor supplies regardless of paradigm. A fetch-time \emph{summary} is exactly that: \VisitSummary{} improves on \VisitSnippet{}'s raw window at every backbone and is \FtE{}'s closest competitor there. What it cannot do is revisit its own rendering, since each page is summarized once and a fact the paraphrase drops is unrecoverable. BrowseComp turns on a single hard-won fact, where what matters is whether a first extraction that misses can be retried rather than the amount of text pulled, and that is the axis along which the paradigms actually differ (§\ref{sec:exp-revisit}).

\paragraph{\FtE{} improves over browsing.} It also beats \OpenFind{} in every cell, by $+2.0$ to $+4.5$\,pp on BrowseComp and ahead on WideSearch Row-F1 throughout. The two are otherwise closely matched: both keep extraction on-demand and read only a window of a page at a time, so extraction is held constant and their gap is the value of the \emph{workspace} itself. That workspace differs from a session in two ways at once. \OpenFind{}'s page is released on the next \tool{open}, whereas every body \FtE{} fetches stays queryable off-context for the rest of the trajectory; and a \tool{find} reaches only the page in hand, whereas \tool{grep} reaches everything held. §\ref{sec:exp-ablation} takes the second away while keeping the first, and finds that retention does not stand on its own: it pays only when the retained pages can be addressed together.

\paragraph{Where the tool budget goes.} The \emph{\#\textit{Search}} column moves against the quality of the document interface, at both ends of the table. \SnippetOnly{}, which has no way to work a page at all, issues the most queries of any profile at every backbone, $70$--$101$ on BrowseComp and $39$--$88$ on WideSearch, against $58$--$81$ and $20$--$62$ for the profiles that can open one; and in all six cells the two on-demand profiles search less than the fetch-time visit family. An agent that cannot get what it needs out of a page it already holds asks the search engine again instead. The ablations confirm the trade from the other direction: every capability we remove from \FtE{} raises \tool{search} by $6$ to $8.3$ calls per question on BrowseComp (§\ref{sec:exp-ablation}). Document access and search are therefore substitutes rather than additive costs, which is also why \FtE{}'s margins cannot be read as the effect of a heavier search system: it is among the two lightest searchers in every cell. Nor does it spend the budget on selection, since the \emph{\#\textit{Select}}/\emph{\#\textit{Extract}} split shows it materializes no more documents than the visit family and often fewer. What it adds is extraction, the \tool{grep} and \tool{read} calls the visit family folds into selection and cannot repeat: cheap selection paired with deferred, repeatable extraction rather than heavier fetching.

\begin{table}[t]
\centering
\small
\caption{\FtE{} tool-call distribution (BC = BrowseComp, WS = WideSearch): mean per-question calls, end-of-trajectory cache size ($\mathcal{C}_q$), the \tool{grep}\,:\,\tool{read} ratio (g\,:\,r), and the share of questions that issue at least one whole-cache \tool{grep} (x-c.\,\%). Q-35B, Q-122B, and DS-V4 are the three backbones of Table~\ref{tab:main}.}
\label{tab:cache-tool-calls}
\setlength{\tabcolsep}{2pt}
\begin{tabular}{ll cccc ccc}
\toprule
& & \multicolumn{4}{c}{Calls / question} & \multicolumn{3}{c}{Derived} \\
\cmidrule(lr){3-6} \cmidrule(lr){7-9}
Model & & \tool{search} & \tool{fetch} & \tool{grep} & \tool{read} & $\mathcal{C}_q$ & g\,:\,r & x-c.\,\% \\
\midrule
\multirow{2}{*}{Q-35B}
  & BC & 58.3 & 5.4  & 3.0  & 3.7 & 5.2  & 0.79 & 7.5 \\
  & WS & 21.6 & 11.2 & 4.6  & 7.4 & 10.9 & 0.62 & 4.0 \\
\midrule
\multirow{2}{*}{Q-122B}
  & BC & 62.8 & 7.3  & 3.7  & 4.1 & 7.0  & 0.89 & 2.5 \\
  & WS & 20.3 & 11.3 & 6.1  & 6.3 & 11.1 & 0.97 & 10.5 \\
\midrule
\multirow{2}{*}{DS-V4}
  & BC & 66.9 & 10.7 & 5.7  & 5.8 & 10.6 & 0.98 & 1.0 \\
  & WS & 36.0 & 15.3 & 12.8 & 8.2 & 15.1 & 1.56 & 4.5 \\
\bottomrule
\end{tabular}
\end{table}

\subsection{Analysis}
\label{sec:exp-analysis}

\paragraph{How \FtE{} uses its tools.}
\label{sec:exp-tool-calls}
Table~\ref{tab:cache-tool-calls} reports \FtE{}'s per-question tool-call counts across all three backbones and both open-web benchmarks; parameter- and sequence-level distributions are in Appendix~\ref{app:supp}. Four readings carry the analysis. First, the benchmarks elicit two regimes: on BrowseComp \tool{search} dominates the budget at a \tool{search}\,:\,\tool{fetch} ratio of $6$--$11$:$1$, whereas on WideSearch it falls to about half of all calls as \tool{fetch}, \tool{grep}, and \tool{read} rise to meet the per-field extraction the task demands. Second, the \tool{grep}\,:\,\tool{read} mix tracks backbone strength: the strongest backbone locates by pattern and reads a small window around the hit, at a ratio near or above $1$:$1$ (up to $1.56$ on WideSearch), while the weakest leans on \tool{read}, down to $0.62$, paging through larger windows to offset coarser localization. Third, the workspace these calls build is modest but real: the end-of-trajectory cache $\mathcal{C}_q$ holds $5$--$11$ documents on BrowseComp and $11$--$15$ on WideSearch, growing with backbone strength on both, and tracks the \tool{fetch} count almost exactly (few pages are re-fetched), so the agent accumulates a standing set of pages and returns to them by re-querying the cache rather than re-materializing. Fourth, cross-cache \tool{grep} is rare at the question level: only $1$--$10.5\%$ of questions ever issue a whole-cache \tool{grep}, and the great majority name a single file. The cross-page reach is thus available but seldom exercised by untrained backbones, which invites the reading that it contributes little; §\ref{sec:exp-ablation} tests that reading and finds the opposite.

\begin{table}[t]
\centering
\small
\caption{Revisit rate (\%) on the two open-web benchmarks across three backbones: the share of materialized documents the agent returns to after leaving. BC = BrowseComp, WS = WideSearch. \textbf{Bold}: highest per backbone/benchmark.}
\label{tab:revisit}
\setlength{\tabcolsep}{4pt}
\begin{tabular}{ll cc}
\toprule
Backbone & Profile & BC & WS \\
\midrule
\multirow{4}{*}{Qwen3.5-35B-A3B}
  & \VisitSnippet{}  & 6.7  & 8.0  \\
  & \VisitSummary{}  & 10.3 & 7.2  \\
  & \OpenFind{}      & 3.8  & 8.6  \\
  \rowcolor{rowhl}& \FtE{} & \textbf{12.6} & \textbf{71.2} \\
\midrule
\multirow{4}{*}{Qwen3.5-122B-A10B}
  & \VisitSnippet{}  & 8.6  & 5.4  \\
  & \VisitSummary{}  & 6.2  & 3.9  \\
  & \OpenFind{}      & 3.3  & 4.4  \\
  \rowcolor{rowhl}& \FtE{} & \textbf{12.4} & \textbf{67.4} \\
\midrule
\multirow{4}{*}{DeepSeek-V4-Pro}
  & \VisitSnippet{}  & 3.0  & 4.5  \\
  & \VisitSummary{}  & 6.3  & 6.0  \\
  & \OpenFind{}      & 3.2  & 7.2  \\
  \rowcolor{rowhl}& \FtE{} & \textbf{19.7} & \textbf{75.6} \\
\bottomrule
\end{tabular}
\end{table}

\paragraph{Revisiting a page after leaving it.}
\label{sec:exp-revisit}
The behavior that most sharply separates a persistent workspace from a transient one is \emph{returning to a page after moving on}. For the transient interfaces, coming back costs a fresh re-\tool{visit}/re-\tool{open}; for \FtE{} the page stays cached and is re-queried by \tool{grep}/\tool{read} at zero materialization cost. We measure the \emph{revisit rate}: the share of materialized documents the agent extracts from again after touching a different document in between (Table~\ref{tab:revisit}). On BrowseComp, \FtE{} revisits $12$--$20\%$ of its documents against $3$--$10\%$ for every transient interface. On WideSearch, whose broad tables force the agent to return to many pages as new fields come due, the gap widens to $8$--$12\times$ at every backbone, with \FtE{} revisiting $67$--$76\%$ of documents while the transient interfaces stay below $9\%$. The two benchmarks separate this behavior from its payoff. WideSearch is where the behavioral gap is widest, but it is also where a faithful fetch-time summary already reaches high coverage on its own (§\ref{sec:exp-main}), so returning more often buys little end-task margin there. BrowseComp shows the narrower behavioral gap and the larger margin in Table~\ref{tab:main}: each question turns on a single fact, so one page worked a second time can decide the answer. What persistence supplies is thus the \emph{opportunity} to return: a page missed or only partially used on a first pass stays queryable off-context for the rest of the trajectory, so an incomplete first hypothesis is correctable many turns later rather than terminal. On the closed corpus, document-level supervision confirms that the returned-to pages are the \emph{gold} ones and the dissociation holds at line resolution (Appendix~\ref{app:triage}).

\begin{table*}[t]
\centering
\small
\caption{Extraction-tool ablation of \FtE{} on the two open-web benchmarks (Qwen3.5-35B-A3B). \FtESingleFile{} confines \tool{grep} to one already-fetched page; \FtENoGrep{} drops \tool{grep}, \FtENoRead{} drops \tool{read}; the persistent cache and \tool{fetch}/\tool{list\_fetched} are unchanged. Columns match Table~\ref{tab:main}, splitting extraction into \tool{grep} / \tool{read} (\tool{fetch} is selection). ``\textrm{---}'': tool not in the toolset. The highlighted \FtE{} (full) row repeats Table~\ref{tab:main}; \textbf{bold} marks the best per column for \emph{Acc}/F1 and \#\textit{Search} (fewest \tool{search} calls).}
\label{tab:ablation}
\setlength{\tabcolsep}{4pt}
\begin{tabular}{l ccccc cccccc}
\toprule
& \multicolumn{5}{c}{\textbf{BrowseComp}} & \multicolumn{6}{c}{\textbf{WideSearch}} \\
\cmidrule(lr){2-6} \cmidrule(lr){7-12}
Profile & Acc & \#\textit{Search} & \tool{fetch} & \tool{grep} & \tool{read} & Row-F1 & Item-F1 & \#\textit{Search} & \tool{fetch} & \tool{grep} & \tool{read} \\
\midrule
\rowcolor{rowhl}\FtE{} (full) & \textbf{44.5} & \textbf{58.3} & 5.4 & 3.0 & 3.7 & \textbf{34.3} & \textbf{63.9} & \textbf{21.6} & 11.2 & 4.6 & 7.4 \\
\midrule
\FtESingleFile{} & 39.5 & 66.6 & 6.5 & 3.5 & 4.3 & 29.2 & 58.2 & 23.0 & 11.1 & 4.5 & 7.8 \\
\FtENoGrep{}    & 37.0 & 65.8 & 4.8 & --- & 6.7 & 33.0 & 62.9 & 22.7 & 9.4 & --- & 10.4 \\
\FtENoRead{}    & 39.5 & 64.4 & 5.8 & 6.6 & --- & 33.9 & 62.2 & 26.1 & 13.1 & 13.1 & --- \\
\bottomrule
\end{tabular}
\end{table*}

\paragraph{Ablating the extraction tools.}
\label{sec:exp-ablation}
\FtE{} gives the agent three capabilities over a retained body: \tool{grep} to locate a hit, \tool{read} to pull a bounded window around it, and a query scope that spans the whole cache. We remove one at a time, on the two open-web benchmarks and the primary Qwen3.5-35B-A3B backbone, the tier where coarser localization stresses the interface most (Table~\ref{tab:ablation}). The cache and \tool{fetch}/\tool{list\_fetched} are untouched throughout, so each arm isolates the dropped capability: without \tool{grep} the agent must find a range by paging rather than by pattern, without \tool{read} it must read through \tool{grep}'s $\pm$context window, and with \tool{grep} confined to one file it must work the cache a page at a time.

\emph{Both tools contribute, and their roles do not collapse into each other.} Dropping either lowers end-task quality, sharply on BrowseComp ($-7.5$\,pp \emph{Acc} without \tool{grep}, $-5.0$ without \tool{read}) and mildly on WideSearch ($\le 1.3$ Row-F1). The traces show the surviving tool taking over the missing one's job, with total extraction flat: without the locator the agent roughly doubles \tool{read}, paging to find ranges it can no longer pattern-match, and without the windowed reader it more than doubles \tool{grep}, widening the $\pm$context window until one call does the work of a pair (the share of \tool{grep} calls asking for eight or more context lines rises threefold on BrowseComp and more than eightfold on WideSearch). Neither substitution is free, and on BrowseComp losing \tool{grep} costs more: an agent without \tool{read} still has a precise locator that doubles as a reader, whereas one without \tool{grep} falls back on blind paging, so localization, not window size, is the scarcer capability.

\emph{The workspace pays only when it is jointly addressable.} Confining \tool{grep} to a single already-fetched page (\FtESingleFile{}) costs as much as dropping an extraction tool outright, $-5.0$\,pp on BrowseComp and $-5.1$ Row-F1 on WideSearch, and the budget goes somewhere revealing. Extraction barely moves, with \tool{grep} and \tool{read} within $0.6$ calls of the full profile, while on BrowseComp \tool{search} rises by $8.3$ calls per question and \tool{fetch} by $1.1$. Stripped of a query that spans the workspace, the agent falls back on the search engine rather than on the pages it already holds: the substitution of §\ref{sec:exp-main}, here produced causally by removing one capability from one profile. The fallback is far weaker on WideSearch, at $1.4$ and $2.0$ more \tool{search} calls, consistent with per-field questions being answerable one page at a time.

This arm also fills in the design space of Table~\ref{tab:profiles}: \FtESingleFile{} retains bodies across the trajectory but addresses them one page at a time, the rung between \OpenFind{} and \FtE{}. It does not sit between them in accuracy. On BrowseComp it falls \emph{below} the transient \OpenFind{} at both backbones, $39.5$ against $42.0$ and $45.5$ against $50.5$: a workspace the agent must address one file at a time carries the bookkeeping of many pages without the reach that makes holding them worthwhile. Persistence and cache-wide scope are therefore two halves of one mechanism rather than additive contributions, which is why we read the \FtE{}--\OpenFind{} contrast as the value of the workspace as a whole. Since the constraint is enforced by refusing cache-wide patterns, part of the drop is the retry cost of a refused call, and this arm is a lower bound on what retention alone delivers.

\section{Conclusion}
\label{sec:conclusion}

We argued that the dominant \textit{visit-and-read} interface fuses document selection with evidence extraction, fixing a rendering before the agent knows what it will need, while \textit{browsing} decouples the two but holds the selected body only in a transient single-page session. Fetch-then-Explore addresses both: pages are fetched into a per-question filesystem workspace and queried later through bounded tools such as \tool{grep} and \tool{read}, rather than entering the context wholesale. It leads BrowseComp accuracy at all three backbones and generally matches or exceeds the baselines on WideSearch. Behind these gains is the move a persistent workspace enables: returning to a page after leaving it, which Fetch-then-Explore does far more than any transient interface, up to $8$--$12\times$ on WideSearch, so a page missed on a first pass stays correctable many turns later rather than terminal.
Two extensions follow: layering cross-trajectory memory on the same substrate, and enriching the cache with fetch-time entity and structure indexes to surface candidates that lexical \tool{grep} cannot. More broadly, decoupling an operation from the medium that holds its output raises a structural question: where an agent's working memory should live.

\medskip

{
\small

\bibliographystyle{plainnat}
\bibliography{ref, search_agent}

@inproceedings{yao2023react,
  title     = {{ReAct}: Synergizing Reasoning and Acting in Language Models},
  author    = {Yao, Shunyu and Zhao, Jeffrey and Yu, Dian and Du, Nan and Shafran, Izhak and Narasimhan, Karthik and Cao, Yuan},
  booktitle = {The Eleventh International Conference on Learning Representations (ICLR)},
  year      = {2023}
}

@misc{lu2026DrDCI,
  title  = {{{Dr-DCI}}: {{Scaling Direct Corpus Interaction}} via {{Dynamic Workspace Expansion}}},
  author = {Lu, Yi and Li, Zhuofeng and Nie, Ping and Zhang, Haoxiang and Zhang, Yuyu and Zou, Kai and Chen, Wenhu and Lin, Jimmy and Jiang, Dongfu and Zhang, Yu},
  year   = {2026},
  eprint = {2606.14885},
  archivePrefix = {arXiv},
  doi    = {10.48550/arXiv.2606.14885},
}

@misc{qwen3.5,
  title  = {{Qwen3.5}: {Towards Native Multimodal Agents}},
  author = {{Qwen Team}},
  month  = {February},
  year   = {2026},
  url    = {https://qwen.ai/blog?id=qwen3.5}
}

@misc{deepseekai2026deepseekv4,
  title  = {{DeepSeek-V4}: {Towards Highly Efficient Million-Token Context Intelligence}},
  author = {{DeepSeek-AI}},
  year   = {2026},
}

@inproceedings{lewis2020rag,
  title     = {Retrieval-Augmented Generation for Knowledge-Intensive {NLP} Tasks},
  author    = {Lewis, Patrick and Perez, Ethan and Piktus, Aleksandra and Petroni, Fabio and Karpukhin, Vladimir and Goyal, Naman and K{\"u}ttler, Heinrich and Lewis, Mike and Yih, Wen-tau and Rockt{\"a}schel, Tim and Riedel, Sebastian and Kiela, Douwe},
  booktitle = {Advances in Neural Information Processing Systems (NeurIPS)},
  year      = {2020},
}

@article{nakano2021webgpt,
  title   = {{WebGPT}: Browser-Assisted Question-Answering with Human Feedback},
  author  = {Nakano, Reiichiro and Hilton, Jacob and Balaji, Suchir and Wu, Jeff and Ouyang, Long and Kim, Christina and Hesse, Christopher and Jain, Shantanu and Kosaraju, Vineet and Saunders, William and Jiang, Xu and Cobbe, Karl and Eloundou, Tyna and Krueger, Gretchen and Button, Kevin and Knight, Matthew and Chess, Benjamin and Schulman, John},
  journal = {arXiv preprint arXiv:2112.09332},
  year    = {2021},
}

@inproceedings{schick2023toolformer,
  title     = {Toolformer: Language Models Can Teach Themselves to Use Tools},
  author    = {Schick, Timo and Dwivedi-Yu, Jane and Dess{\`i}, Roberto and Raileanu, Roberta and Lomeli, Maria and Hambro, Eric and Zettlemoyer, Luke and Cancedda, Nicola and Scialom, Thomas},
  booktitle = {Advances in Neural Information Processing Systems (NeurIPS)},
  year      = {2023},
}

@inproceedings{press2023selfask,
  title     = {Measuring and Narrowing the Compositionality Gap in Language Models},
  author    = {Press, Ofir and Zhang, Muru and Min, Sewon and Schmidt, Ludwig and Smith, Noah A. and Lewis, Mike},
  booktitle = {Findings of the Association for Computational Linguistics: EMNLP},
  year      = {2023},
}

@inproceedings{trivedi2023ircot,
  title     = {Interleaving Retrieval with Chain-of-Thought Reasoning for Knowledge-Intensive Multi-Step Questions},
  author    = {Trivedi, Harsh and Balasubramanian, Niranjan and Khot, Tushar and Sabharwal, Ashish},
  booktitle = {Proceedings of the 61st Annual Meeting of the Association for Computational Linguistics (ACL)},
  year      = {2023},
}

@inproceedings{voorhees1999trec8,
  title     = {The {TREC-8} Question Answering Track Report},
  author    = {Voorhees, Ellen M.},
  booktitle = {Proceedings of the Eighth Text REtrieval Conference (TREC-8)},
  year      = {1999},
}

@inproceedings{moldovan2000structure,
  title     = {The Structure and Performance of an Open-Domain Question Answering System},
  author    = {Moldovan, Dan and Harabagiu, Sanda and Pasca, Marius and Mihalcea, Rada and Girju, Roxana and Goodrum, Richard and Rus, Vasile},
  booktitle = {Proceedings of the 38th Annual Meeting of the Association for Computational Linguistics (ACL)},
  year      = {2000},
}

@inproceedings{prager2000predictive,
  title     = {Question-Answering by Predictive Annotation},
  author    = {Prager, John and Brown, Eric and Coden, Anni and Radev, Dragomir},
  booktitle = {Proceedings of the 23rd Annual International ACM SIGIR Conference on Research and Development in Information Retrieval (SIGIR)},
  year      = {2000},
}

@inproceedings{tellex2003passage,
  title     = {Quantitative Evaluation of Passage Retrieval Algorithms for Question Answering},
  author    = {Tellex, Stefanie and Katz, Boris and Lin, Jimmy and Fernandes, Aaron and Marton, Gregory},
  booktitle = {Proceedings of the 26th Annual International ACM SIGIR Conference on Research and Development in Information Retrieval (SIGIR)},
  year      = {2003},
}

@inproceedings{ahn2004wikipedia,
  title     = {Using {Wikipedia} at the {TREC} {QA} Track},
  author    = {Ahn, David and Jijkoun, Valentin and Mishne, Gilad and M{\"u}ller, Karin and de Rijke, Maarten and Schlobach, Stefan},
  booktitle = {Proceedings of the Thirteenth Text REtrieval Conference (TREC 2004)},
  year      = {2004},
}

@inproceedings{chen2017drqa,
  title     = {Reading {Wikipedia} to Answer Open-Domain Questions},
  author    = {Chen, Danqi and Fisch, Adam and Weston, Jason and Bordes, Antoine},
  booktitle = {Proceedings of the 55th Annual Meeting of the Association for Computational Linguistics (ACL)},
  year      = {2017},
}

@article{gpt-oss,
  title={gpt-oss-120b \& gpt-oss-20b model card},
  author={Agarwal, Sandhini and Ahmad, Lama and Ai, Jason and Altman, Sam and Applebaum, Andy and Arbus, Edwin and Arora, Rahul K and Bai, Yu and Baker, Bowen and Bao, Haiming and others},
  journal={arXiv preprint arXiv:2508.10925},
  year={2025}
}

@inproceedings{yang2024sweagent,
  title     = {{SWE-agent}: Agent-Computer Interfaces Enable Automated Software Engineering},
  author    = {Yang, John and Jimenez, Carlos E. and Wettig, Alexander and Lieret, Kilian and Yao, Shunyu and Narasimhan, Karthik and Press, Ofir},
  booktitle = {Advances in Neural Information Processing Systems (NeurIPS)},
  year      = {2024}
}

@inproceedings{wang2024openhands,
  title     = {{OpenHands}: An Open Platform for {AI} Software Developers as Generalist Agents},
  author    = {Wang, Xingyao and Li, Boxuan and Song, Yufan and Xu, Frank F. and Tang, Xiangru and Zhuge, Mingchen and Pan, Jiayi and Song, Yueqi and Li, Bowen and Singh, Jaskirat and Tran, Hoang H. and Li, Fuqiang and Ma, Ren and Zheng, Mingzhang and Qian, Bill and Shao, Yanjun and Muennighoff, Niklas and Zhang, Yizhe and Hui, Binyuan and Lin, Junyang and Brennan, Robert and Peng, Hao and Ji, Heng and Neubig, Graham},
  booktitle = {The Thirteenth International Conference on Learning Representations (ICLR)},
  year      = {2025}
}

@misc{Jiang2026Harness-1ReinforcementLearning,
  title = {Harness-1: Reinforcement Learning for Search Agents with State-Externalizing Harnesses},
  author = {Pengcheng Jiang and Zhiyi Shi and Kelly Hong and Xueqiang Xu and Jiashuo Sun and Jimeng Sun and Hammad Bashir and Jiawei Han},
  year = {2026},
  archivePrefix = {arXiv},
  eprint = {2606.02373},
  url = {https://arxiv.org/abs/2606.02373}
}

@misc{Salemi2026GrepSeekTrainingSearch,
  title = {GrepSeek: Training Search Agents for Direct Corpus Interaction},
  author = {Alireza Salemi and Chang Zeng and Atharva Nijasure and Jui-Hui Chung and Razieh Rahimi and Fernando Diaz and Hamed Zamani},
  year = {2026},
  archivePrefix = {arXiv},
  eprint = {2605.29307},
  url = {https://arxiv.org/abs/2605.29307}
}

@misc{Fan2026LiveBrowseCompAreSearch,
  title = {LiveBrowseComp: Are Search Agents Searching, or Just Verifying What They Already Know?},
  author = {HuiMing Fan and Xiao Wang and Zheng Chu and Qianyu Wang and Zhuoyao Wang and Ming Liu and Bing Qin and XingYu},
  year = {2026},
  archivePrefix = {arXiv},
  eprint = {2605.28721},
  url = {https://arxiv.org/abs/2605.28721}
}

@misc{Sen2026IsGrepAll,
  title = {Is Grep All You Need? How Agent Harnesses Reshape Agentic Search},
  author = {Sahil Sen and Akhil Kasturi and Elias Lumer and Anmol Gulati and Vamse Kumar Subbiah},
  year = {2026},
  archivePrefix = {arXiv},
  eprint = {2605.15184},
  url = {https://arxiv.org/abs/2605.15184}
}

@misc{Wang2026EvoBrowseCompBenchmarkingSearch,
  title = {EvoBrowseComp: Benchmarking Search Agents on Evolving Knowledge},
  author = {Yunhan Wang and Jiaan Wang and Lianzhe Huang and Xianfeng Zeng and Fandong Meng},
  year = {2026},
  archivePrefix = {arXiv},
  eprint = {2606.13120},
  url = {https://arxiv.org/abs/2606.13120}
}

@misc{Zhuang2026TowardsRetrievingInteraction,
  title = {Towards Retrieving Interaction Spaces for Agentic Search},
  author = {Shengyao Zhuang and Yuansheng Ni and Hengxin Fun and Jimmy Lin and Xueguang Ma},
  year = {2026},
  archivePrefix = {arXiv},
  eprint = {2606.06880},
  url = {https://arxiv.org/abs/2606.06880}
}

@misc{Li2026BeyondSemanticSimilarity,
  title = {Beyond Semantic Similarity: Rethinking Retrieval for Agentic Search via Direct Corpus Interaction},
  author = {Zhuofeng Li and Haoxiang Zhang and Cong Wei and Pan Lu and Ping Nie and Yi Lu and Yuyang Bai and Shangbin Feng and Hangxiao Zhu and Ming Zhong and Yuyu Zhang and Jianwen Xie and Yejin Choi and James Zou and Jiawei Han and Wenhu Chen and Jimmy Lin and Dongfu Jiang and Yu Zhang},
  year = {2026},
  archivePrefix = {arXiv},
  eprint = {2605.05242},
  url = {https://arxiv.org/abs/2605.05242}
}

@misc{Zhu2026FS-ResearcherTest-TimeScaling,
  title = {FS-Researcher: Test-Time Scaling for Long-Horizon Research Tasks with File-System-Based Agents},
  author = {Chiwei Zhu and Benfeng Xu and Mingxuan Du and Shaohan Wang and Xiaorui Wang and Zhendong Mao and Yongdong Zhang},
  year = {2026},
  archivePrefix = {arXiv},
  eprint = {2602.01566},
  url = {https://arxiv.org/abs/2602.01566}
}

@misc{Zhou2025BrowseComp-ZHBenchmarkingWeb,
  title = {BrowseComp-ZH: Benchmarking Web Browsing Ability of Large Language Models in Chinese},
  author = {Peilin Zhou and Bruce Leon and Xiang Ying and Can Zhang and Yifan Shao and Qichen Ye and Dading Chong and Zhiling Jin and Chenxuan Xie and Meng Cao and Yuxin Gu and Sixin Hong and Jing Ren and Jian Chen and Chao Liu and Yining Hua},
  year = {2025},
  archivePrefix = {arXiv},
  eprint = {2504.19314},
  url = {https://arxiv.org/abs/2504.19314}
}

@misc{Zheng2025DeepResearcherScalingDeep,
  title = {DeepResearcher: Scaling Deep Research via Reinforcement Learning in Real-world Environments},
  author = {Yuxiang Zheng and Dayuan Fu and Xiangkun Hu and Xiaojie Cai and Lyumanshan Ye and Pengrui Lu and Pengfei Liu},
  year = {2025},
  archivePrefix = {arXiv},
  eprint = {2504.03160},
  url = {https://arxiv.org/abs/2504.03160}
}

@misc{Ye2025AgentFoldLong-HorizonWeb,
  title = {AgentFold: Long-Horizon Web Agents with Proactive Context Management},
  author = {Rui Ye and Zhongwang Zhang and Kuan Li and Huifeng Yin and Zhengwei Tao and Yida Zhao and Liangcai Su and Liwen Zhang and Zile Qiao and Xinyu Wang and Pengjun Xie and Fei Huang and Siheng Chen and Jingren Zhou and Yong Jiang},
  year = {2025},
  archivePrefix = {arXiv},
  eprint = {2510.24699},
  url = {https://arxiv.org/abs/2510.24699}
}

@misc{Yan2025GeneralAgenticMemory,
  title = {General Agentic Memory Via Deep Research},
  author = {B. Y. Yan and Chaofan Li and Hongjin Qian and Shuqi Lu and Zheng Liu},
  year = {2025},
  archivePrefix = {arXiv},
  eprint = {2511.18423},
  url = {https://arxiv.org/abs/2511.18423}
}

@misc{Xi2025ASurveyof,
  title = {A Survey of LLM-based Deep Search Agents: Paradigm, Optimization, Evaluation, and Challenges},
  author = {Yunjia Xi and Jianghao Lin and Yongzhao Xiao and Zheli Zhou and Rong Shan and Te Gao and Jiachen Zhu and Weiwen Liu and Yong Yu and Weinan Zhang},
  year = {2025},
  archivePrefix = {arXiv},
  eprint = {2508.05668},
  url = {https://arxiv.org/abs/2508.05668}
}

@misc{Wu2025WebDancerTowardsAutonomous,
  title = {WebDancer: Towards Autonomous Information Seeking Agency},
  author = {Jialong Wu and Baixuan Li and Runnan Fang and Wenbiao Yin and Liwen Zhang and Zhengwei Tao and Dingchu Zhang and Zekun Xi and Yong Jiang and Pengjun Xie and Fei Huang and Jingren Zhou},
  year = {2025},
  archivePrefix = {arXiv},
  eprint = {2505.22648},
  url = {https://arxiv.org/abs/2505.22648}
}

@misc{Wu2025ReSumUnlockingLong-Horizon,
  title = {ReSum: Unlocking Long-Horizon Search Intelligence via Context Summarization},
  author = {Xixi Wu and Kuan Li and Yida Zhao and Liwen Zhang and Litu Ou and Huifeng Yin and Zhongwang Zhang and Yong Jiang and Pengjun Xie and Fei Huang and Minhao Cheng and Shuai Wang and Hong Cheng and Jingren Zhou},
  year = {2025},
  archivePrefix = {arXiv},
  eprint = {2509.13313},
  url = {https://arxiv.org/abs/2509.13313}
}

@misc{Wong2025WideSearchBenchmarkingAgentic,
  title = {WideSearch: Benchmarking Agentic Broad Info-Seeking},
  author = {Ryan Wong and Jiawei Wang and Junjie Zhao and Li Chen and Yan Gao and Long Zhang and Xuan Zhou and Zuo Wang and Kai Xiang and Ge Zhang and Wenhao Huang and Yang Wang and Ke Wang},
  year = {2025},
  archivePrefix = {arXiv},
  eprint = {2508.07999},
  url = {https://arxiv.org/abs/2508.07999}
}

@misc{Wei2025BrowseCompASimple,
  title = {BrowseComp: A Simple Yet Challenging Benchmark for Browsing Agents},
  author = {Jason Wei and Zhiqing Sun and Spencer Papay and Scott McKinney and Jeffrey Han and Isa Fulford and Hyung Won Chung and Alex Tachard Passos and William Fedus and Amelia Glaese},
  year = {2025},
  archivePrefix = {arXiv},
  eprint = {2504.12516},
  url = {https://arxiv.org/abs/2504.12516}
}

@misc{Team2025TongyiDeepResearchTechnical,
  title = {Tongyi DeepResearch Technical Report},
  author = {Tongyi DeepResearch Team and Baixuan Li and Bo Zhang and Dingchu Zhang and Fei Huang and Guangyu Li and Guoxin Chen and Huifeng Yin and Jialong Wu and Jingren Zhou and Kuan Li and Liangcai Su and Litu Ou and Liwen Zhang and Pengjun Xie and Rui Ye and Wenbiao Yin and Xinmiao Yu and Xinyu Wang and Xixi Wu and Xuanzhong Chen and Yida Zhao and Zhen Zhang and Zhengwei Tao and Zhongwang Zhang and Zile Qiao and Chenxi Wang and Donglei Yu and Gang Fu and Haiyang Shen and Jiayin Yang and Jun Lin and Junkai Zhang and Kui Zeng and Li Yang and Hailong Yin and Maojia Song and Ming Yan and Peng Xia and Qian Xiao and Rui Min and Ruixue Ding and Runnan Fang and Shaowei Chen and Shen Huang and Shihang Wang and Shihao Cai and Weizhou Shen and Xiaobin Wang and Xin Guan and Xinyu Geng and Yingcheng Shi and Yuning Wu and Zhuo Chen and Zijian Li and Yong Jiang},
  year = {2025},
  archivePrefix = {arXiv},
  eprint = {2510.24701},
  url = {https://arxiv.org/abs/2510.24701}
}

@misc{Team2025MiroThinkerPushingthe,
  title = {MiroThinker: Pushing the Performance Boundaries of Open-Source Research Agents via Model, Context, and Interactive Scaling},
  author = {MiroMind Team and Song Bai and Lidong Bing and Carson Chen and Guanzheng Chen and Yuntao Chen and Zhe Chen and Ziyi Chen and Jifeng Dai and Xuan Dong and Wenhan Dou and Yue Deng and Yunjie Fu and Junqi Ge and Chenxia Han and Tammy Huang and Zhenhang Huang and Jerry Jiao and Shilei Jiang and Tianyu Jiao and Xiaoqi Jian and Lei Lei and Ruilin Li and Ryan Luo and Tiantong Li and Xiang Lin and Ziyuan Liu and Zhiqi Li and Jie Ni and Qiang Ren and Pax Sun and Shiqian Su and Chenxin Tao and Bin Wang and Hellen Wang and Haonan Wang and James Wang and Jin Wang and Jojo Wang and Letian Wang and Shizun Wang and Weizhi Wang and Zixuan Wang and Jinfan Xu and Sen Xing and Chenyu Yang and Hai Ye and Jiaheng Yu and Yue Yu and Muyan Zhong and Tianchen Zhao and Xizhou Zhu and Yanpeng Zhou and Yifan Zhang and Zhi Zhu},
  year = {2025},
  archivePrefix = {arXiv},
  eprint = {2511.11793},
  url = {https://arxiv.org/abs/2511.11793}
}

@misc{Su2025ScalingAgentsvia,
  title = {Scaling Agents via Continual Pre-training},
  author = {Liangcai Su and Zhen Zhang and Guangyu Li and Zhuo Chen and Chenxi Wang and Maojia Song and Xinyu Wang and Kuan Li and Jialong Wu and Xuanzhong Chen and Zile Qiao and Zhongwang Zhang and Huifeng Yin and Shihao Cai and Runnan Fang and Zhengwei Tao and Wenbiao Yin and Chenxiong Qian and Yong Jiang and Pengjun Xie and Fei Huang and Jingren Zhou},
  year = {2025},
  archivePrefix = {arXiv},
  eprint = {2509.13310},
  url = {https://arxiv.org/abs/2509.13310}
}

@misc{Shi2025DeepResearchA,
  title = {Deep Research: A Systematic Survey},
  author = {Zhengliang Shi and Yiqun Chen and Haitao Li and Weiwei Sun and Shiyu Ni and Yougang Lyu and Run-Ze Fan and Bowen Jin and Yixuan Weng and Minjun Zhu and Qiujie Xie and Xinyu Guo and Qu Yang and Jiayi Wu and Jujia Zhao and Xiaqiang Tang and Xinbei Ma and Cunxiang Wang and Jiaxin Mao and Qingyao Ai and Jen-Tse Huang and Wenxuan Wang and Yue Zhang and Yiming Yang and Zhaopeng Tu and Zhaochun Ren},
  year = {2025},
  archivePrefix = {arXiv},
  eprint = {2512.02038},
  url = {https://arxiv.org/abs/2512.02038}
}

@misc{Qiao2025WebResearcherUnleashingUnbounded,
  title = {WebResearcher: Unleashing Unbounded Reasoning Capability in Long-Horizon Agents},
  author = {Zile Qiao and Guoxin Chen and Xuanzhong Chen and Donglei Yu and Wenbiao Yin and Xinyu Wang and Zhen Zhang and Baixuan Li and Huifeng Yin and Kuan Li and Rui Min and Minpeng Liao and Yong Jiang and Pengjun Xie and Fei Huang and Jingren Zhou},
  year = {2025},
  archivePrefix = {arXiv},
  eprint = {2509.13309},
  url = {https://arxiv.org/abs/2509.13309}
}

@misc{Li2026OpenResearcherAFully,
  title = {OpenResearcher: A Fully Open Pipeline for Long-Horizon Deep Research Trajectory Synthesis},
  author = {Zhuofeng Li and Dongfu Jiang and Xueguang Ma and Haoxiang Zhang and Ping Nie and Yuyu Zhang and Kai Zou and Jianwen Xie and Yu Zhang and Wenhu Chen},
  year = {2026},
  archivePrefix = {arXiv},
  eprint = {2603.20278},
  url = {https://arxiv.org/abs/2603.20278}
}

@misc{Li2025WebThinkerEmpoweringLarge,
  title = {WebThinker: Empowering Large Reasoning Models with Deep Research Capability},
  author = {Xiaoxi Li and Jiajie Jin and Guanting Dong and Hongjin Qian and Yutao Zhu and Yongkang Wu and Ji-Rong Wen and Zhicheng Dou},
  year = {2025},
  archivePrefix = {arXiv},
  eprint = {2504.21776},
  url = {https://arxiv.org/abs/2504.21776}
}

@misc{Li2025WebSailorNavigatingSuper-human,
  title = {WebSailor: Navigating Super-human Reasoning for Web Agent},
  author = {Kuan Li and Zhongwang Zhang and Huifeng Yin and Liwen Zhang and Litu Ou and Jialong Wu and Wenbiao Yin and Baixuan Li and Zhengwei Tao and Xinyu Wang and Weizhou Shen and Junkai Zhang and Dingchu Zhang and Xixi Wu and Yong Jiang and Ming Yan and Pengjun Xie and Fei Huang and Jingren Zhou},
  year = {2025},
  archivePrefix = {arXiv},
  eprint = {2507.02592},
  url = {https://arxiv.org/abs/2507.02592}
}

@misc{Li2025DeepAgentAGeneral,
  title = {DeepAgent: A General Reasoning Agent with Scalable Toolsets},
  author = {Xiaoxi Li and Wenxiang Jiao and Jiarui Jin and Guanting Dong and Jiajie Jin and Yinuo Wang and Hao Wang and Yutao Zhu and Ji-Rong Wen and Yuan Lu and Zhicheng Dou},
  year = {2025},
  archivePrefix = {arXiv},
  eprint = {2510.21618},
  url = {https://arxiv.org/abs/2510.21618}
}

@misc{Du2026OpenSeekerDemocratizingFrontier,
  title = {OpenSeeker: Democratizing Frontier Search Agents by Fully Open-Sourcing Training Data},
  author = {Yuwen Du and Rui Ye and Shuo Tang and Xinyu Zhu and Yijun Lu and Yuzhu Cai and Siheng Chen},
  year = {2026},
  archivePrefix = {arXiv},
  eprint = {2603.15594},
  url = {https://arxiv.org/abs/2603.15594}
}

@misc{Deng2025InteractCompEvaluatingSearch,
  title = {InteractComp: Evaluating Search Agents With Ambiguous Queries},
  author = {Mingyi Deng and Lijun Huang and Yani Fan and Jiayi Zhang and Fashen Ren and Jinyi Bai and Fuzhen Yang and Dayi Miao and Zhaoyang Yu and Yifan Wu and Yanfei Zhang and Fengwei Teng and Yingjia Wan and Song Hu and Yude Li and Xin Jin and Conghao Hu and Haoyu Li and Qirui Fu and Tai Zhong and Xinyu Wang and Xiangru Tang and Nan Tang and Chenglin Wu and Yuyu Luo},
  year = {2025},
  archivePrefix = {arXiv},
  eprint = {2510.24668},
  url = {https://arxiv.org/abs/2510.24668}
}

@misc{Chen2025IterResearchRethinkingLong-Horizon,
  title = {IterResearch: Rethinking Long-Horizon Agents with Interaction Scaling},
  author = {Guoxin Chen and Zile Qiao and Xuanzhong Chen and Donglei Yu and Haotian Xu and Wayne Xin Zhao and Ruihua Song and Wenbiao Yin and Huifeng Yin and Liwen Zhang and Kuan Li and Minpeng Liao and Yong Jiang and Pengjun Xie and Fei Huang and Jingren Zhou},
  year = {2025},
  archivePrefix = {arXiv},
  eprint = {2511.07327},
  url = {https://arxiv.org/abs/2511.07327}
}

@misc{Chen2025BrowseComp-PlusAMore,
  title = {BrowseComp-Plus: A More Fair and Transparent Evaluation Benchmark of Deep-Research Agent},
  author = {Zijian Chen and Xueguang Ma and Shengyao Zhuang and Ping Nie and Kai Zou and Andrew Liu and Joshua Green and Kshama Patel and Ruoxi Meng and Mingyi Su and Sahel Sharifymoghaddam and Yanxi Li and Haoran Hong and Xinyu Shi and Xuye Liu and Nandan Thakur and Crystina Zhang and Luyu Gao and Wenhu Chen and Jimmy Lin},
  year = {2025},
  archivePrefix = {arXiv},
  eprint = {2508.06600},
  url = {https://arxiv.org/abs/2508.06600}
}

}

\appendix

\section{Implementation Details}
\label{app:impl}

\subsection{Agent Implementation}
\label{app:impl-agent}

\subsubsection{Model backend and sampling parameters}
\label{app:impl-agent-model}

The agent sampling uses $\texttt{temperature}=1.0$, $\texttt{top\_p}=0.95$, $\texttt{max\_tokens}=16{,}384$, and $\texttt{presence\_penalty}=0.0$. Two LLM backends are supported: \texttt{vllm} for self-hosted Qwen models and API for DeepSeek-V4-Pro. Both use the native tool-calling support.

\subsubsection{System prompt}
\label{app:impl-agent-prompt}

The agent receives the following system prompt, assembled from a fixed body, an optional variant-specific addendum, and a date footer:

\begin{promptbox}
You are a deep research assistant. Your core function is to conduct thorough, multi-source investigations into any topic. You must handle both broad, open-domain inquiries and queries within specialized academic fields. For every request, synthesize information from credible, diverse sources to deliver a comprehensive, accurate, and objective response. When you have gathered sufficient information and are ready to provide the definitive response, you must enclose the entire final answer within \texttt{<answer></answer>} tags.

\# Tool usage

Use the available native tools whenever you need fresh evidence.

[optional variant-specific addendum]

Current date: YYYY-MM-DD
\end{promptbox}

No per-profile description of the tools is added to the system prompt; the model learns the tool surface solely from the JSON schemas that the harness transmits as native tool definitions. The system prompt is shared identically across all tool profiles and backbones (§\ref{sec:method-agent}).

\subsection{Tool Definitions and Implementation}
\label{app:impl-tools}

\subsubsection{Tool surface reference}
\label{app:impl-tools-surface}

Table~\ref{tab:tool-surface} lists the full signatures and per-tool output bounds of the Fetch-then-Explore tool surface (§\ref{sec:method-cache}), for the structured-tool form and the unified-shell re-skin.

\begin{table*}[t]
\centering
\small
\caption{The Fetch-then-Explore tool surface (code profile id \texttt{fetch\_explore}). \emph{Resource} is the canonical URL returned by the search service; returned bytes are bounded per tool by the last column.}
\label{tab:tool-surface}
\setlength{\tabcolsep}{4pt}
\begin{tabular}{llp{4.8cm}}
\toprule
Tool & Signature & Returned to context \\
\midrule
\tool{search}& $(\text{query}) \to [\{\text{title}, \text{resource}, \text{snippet}\}]$ & top-$K$ results, snippet fixed-length \\
\tool{fetch}         & $(\text{resource}) \to (\text{path}, \text{characters}, \text{status})$ & path + size header, \emph{never the body} \\
\tool{grep}          & $(\text{pattern}, \text{path}, \text{context}) \to \text{matches}$ & up to \texttt{GREP\_MAX\_MATCHES}\,$= 20$ matches grouped by file; each match carries $\pm\text{context}$ lines around it (default $3$, capped at $10$; \texttt{context}$=0$ gives the matching line alone) \\
\tool{read}          & $(\text{path}, \text{offset}, \text{limit}) \to \text{line range}$ & at most \texttt{READ\_LIMIT}\,$= 200$ lines, capped by a $50$\,KB output ceiling \\
\tool{list\_fetched} & $() \to [(\text{path}, \text{resource}, \text{chars})]$ & one header line per fetched file \\
\bottomrule
\end{tabular}
\end{table*}

\subsubsection{Cache substrate on disk}
\label{app:impl-tools-cache}

Each trajectory receives a private cache root created with \texttt{tempfile.mkdtemp(prefix="work-")}; the root is removed via \texttt{shutil.rmtree} at trajectory end (registered as an \texttt{atexit} hook so crashes still clean up). A fetched resource is written as a single UTF-8 text file under this root. Filenames are deterministic in the resource identifier so that a repeated \tool{fetch} is a no-op: for URL-keyed resources the name is \texttt{\{host\_slug\}-\{16-char~sha256\_prefix\}.txt} (e.g.\ \texttt{example-com-a1b2c3d4e5f6g7h8.txt}), where the safe prefix is the first $\le 32$ alphanumeric/hyphen characters of the docid. Each file begins with a three-line provenance header:
\begin{codeblock}
  URL: <canonical resource URL>
  Fetched-At: <ISO-8601 UTC timestamp>
  Characters: <content_length>
\end{codeblock}
(or \texttt{Docid:} for local-backend files), followed by a blank line and the normalized plain-text body. Because the filename is a hash of the resource identifier, a repeated \tool{fetch} short-circuits: the body is already on disk, and only the header (path, size, newly/already-fetched flag) is returned to the agent --- the body never enters the context.

\subsubsection{Tool Schemas}
\label{app:impl-tools-structured}

The model sees each tool as a native function definition with \texttt{name}, \texttt{description}, and JSON Schema \texttt{parameters}. The exact specifications the harness transmits to the model are reproduced below.

\paragraph{\tool{search}}.

\begin{codeblock}
{
  "type": "function",
  "function": {
    "name": "search",
    "description": "Search the web or local corpus and return ranked
      results for a single query.",
    "parameters": {
      "type": "object",
      "properties": {
        "query": {
          "type": "string",
          "description": "A single search query."
        }
      },
      "required": ["query"]
    }
  }
}
\end{codeblock}

\paragraph{\tool{fetch}}.

\begin{codeblock}
{
  "type": "function",
  "function": {
    "name": "fetch",
    "description": "Download a webpage into the per-question workspace and return only the saved file path. This issues an actual HTTP request to the URL; it is NOT a lookup in a pre-existing index. The full content is written to disk — use grep to search across fetched files, read to inspect bounded slices, or bash for shell commands. Content is deliberately NOT returned inline to keep context manageable. Check list_fetched to see what is already in the workspace.",
    "parameters": {
      "type": "object",
      "properties": {
        "url": {
          "type": "string",
          "description": "A single URL to download."
        }
      },
      "required": ["url"]
    }
  }
}
\end{codeblock}

\paragraph{\tool{grep}}.

\begin{codeblock}
{
  "type": "function",
  "function": {
    "name": "grep",
    "description": "Search fetched files for regex pattern matches. By default returns the matching lines only (no surrounding context); pass context=N to include $pm$N lines around each match (like grep -C) so you can read a hit in place instead of a follow-up read. The pattern is a case-insensitive regex, so pass a plain keyword or a regex (e.g. born|birth to match any of several terms). By default (omit the path argument), grep searches across ALL fetched files at once — the recommended way to find which page(s) mention an entity when you have fetched several URLs and do not yet know which one carries the answer. Pass an explicit path only when you already know which single file to scan. Output is grouped by file: each file with matches gets a header line <path>:, followed by   N: text entries (N is the 1-indexed line number, text is the matched line). Use the line number N directly as offset in a follow-up read(path, offset=N) call. Examples: grep(pattern='Marie Curie') searches every fetched file; grep(pattern='born', context=2) shows each hit with 2 lines on each side; grep(pattern='born in', path='/tmp/work-XXXX/foo.txt') scans just that one file.",
    "parameters": {
      "type": "object",
      "properties": {
        "pattern": {
          "type": "string",
          "description": "A plain keyword or a case-insensitive regex
            (e.g. Marie|Curie)."
        },
        "path": {
          "type": "string",
          "description": "Optional. File path or glob pattern. If omitted, grep searches ALL fetched files at once (recommended default). Specify a single path only to narrow an already-located match."
        },
        "context": {
          "type": "integer",
          "description": "Optional. Number of surrounding lines to show on each side of a match (like grep -C). Omit to get a small default context; pass 0 for matching lines only, or a larger value (up to 10) to read a match in place without a separate read."
        }
      },
      "required": ["pattern"]
    }
  }
}
\end{codeblock}

\paragraph{\tool{read}}.

\begin{codeblock}
{
  "type": "function",
  "function": {
    "name": "read",
    "description": "Read a bounded LINE range from a specific fetched file (as returned by fetch or list_fetched). offset is the 1-indexed line number to start reading from (default 1); limit is the number of lines to return per call. Each output line is prefixed with its line number (N: text), so the line numbers grep returns plug straight in as offset here. The trailing message tells you exactly what to pass to continue: (Showing lines X-Y of Z. Use offset=N to continue.). Total output is also capped at ~50 KB to keep observations bounded, regardless of limit.",
    "parameters": {
      "type": "object",
      "properties": {
        "path": {
          "type": "string",
          "description": "File path returned by fetch or list_fetched."
        },
        "offset": {
          "type": "integer",
          "description": "Optional. 1-indexed line number to start reading from. Defaults to 1. Pass the line number returned by grep to jump straight to a match."
        },
        "limit": {
          "type": "integer",
          "description": "Optional. Maximum number of lines to return. Capped by a configured ceiling; a 50 KB output-byte cap also applies, so very long lines may shorten the effective window."
        }
      },
      "required": ["path"]
    }
  }
}
\end{codeblock}

\paragraph{\tool{list\_fetched}}.

\begin{codeblock}
{
  "type": "function",
  "function": {
    "name": "list_fetched",
    "description": "List all documents currently fetched for this question with their file paths and source identifiers. Use to discover available files before grep or read.",
    "parameters": {
      "type": "object",
      "properties": {}
    }
  }
}
\end{codeblock}

\paragraph{\tool{visit} (baseline)} The visit behaviour depends on the tool profile (snippet / summary / full); the caller does not need to know which mode is active.

\begin{codeblock}
{
  "type": "function",
  "function": {
    "name": "visit",
    "description": "Visit a single webpage and summarize information relevant to a specific goal.",
    "parameters": {
      "type": "object",
      "properties": {
        "url": {
          "type": "string",
          "description": "A single URL to visit."
        },
        "goal": {
          "type": "string",
          "description": "The information goal for the page visit."
        }
      },
      "required": ["url", "goal"]
    }
  }
}
\end{codeblock}

\paragraph{\tool{open}} (baseline).

\begin{codeblock}
{
  "type": "function",
  "function": {
    "name": "open",
    "description": "Open a webpage and return its extracted text (truncated). Subsequent find(pattern) calls search within THIS page. Open a different page to switch context.",
    "parameters": {
      "type": "object",
      "properties": {
        "url": {
          "type": "string",
          "description": "A single URL to open."
        }
      },
      "required": ["url"]
    }
  }
}
\end{codeblock}

\paragraph{\tool{find}} (baseline).

\begin{codeblock}
{
  "type": "function",
  "function": {
    "name": "find",
    "description": "Search the currently open page (from a prior open call) for regex or literal text matches. Returns matching lines prefixed with line numbers. Use open() to load a page first, then find() to locate specific information inside it.",
    "parameters": {
      "type": "object",
      "properties": {
        "pattern": {
          "type": "string",
          "description": "Pattern to find in the current page."
        }
      },
      "required": ["pattern"]
    }
  }
}
\end{codeblock}

\subsection{Evaluation Implementation}
\label{app:impl-eval}

All three benchmarks are evaluated using the official LLM-as-a-judge protocol provided by each benchmark. For BrowseComp and BrowseComp-Plus, the judge is DeepSeek-V4-Pro and outputs a structured response with fields \texttt{extracted\_final\_answer}, \texttt{reasoning}, \texttt{correct} (yes/no), and \texttt{confidence} ($0$--$100$); a prediction is scored Correct iff \texttt{correct=yes}. For WideSearch, the judge follows the benchmark's official two-step protocol: a vocabulary-alignment step maps the agent's table to the gold schema, followed by an LLM-judge column step that scores each field. The final Acc is the strict all-or-nothing full-table match; Row-F1 and Item-F1 are computed over the aligned tables by the official scorer. The same judge configuration is applied identically to every profile and backbone, so judge choice enters all cells equally; our claims rest on \emph{intra-backbone} deltas (each comparison in Table~\ref{tab:main} is within one fixed backbone), which a uniform judge bias cannot manufacture or erase. The judge shares its model with one of the three agent backbones, so the DeepSeek-V4-Pro rows are scored by the same model that produced them. Any self-preference this induces applies to all five profiles in that row alike, and so shifts the row's absolute level rather than the intra-backbone deltas we read; the two Qwen backbones are judged by a different model family throughout.

\section{Supplementary Experimental Results}
\label{app:supp}

The per-question tool-call counts of \FtE{} are already cross-backbone in the main text (Table~\ref{tab:cache-tool-calls}). This section adds the parameter-level, call-sequence, and fetch-reuse distributions referenced in §\ref{sec:exp-tool-calls}, all on \FtE{} across the two open-web benchmarks (BC = BrowseComp, WS = WideSearch) and all three backbones. The two-level triage analysis (§\ref{sec:exp-triage}) runs on BrowseComp-Plus, which we evaluate on Qwen3.5-35B-A3B only; it is single-backbone by construction.

\paragraph{\tool{grep} pattern shape and scope}
\label{app:supp-grep}

Table~\ref{tab:app-grep} reports the \tool{grep} pattern-length distribution (in characters) and the path-argument breakdown. Two qualitative facts hold across all runs: $74$--$92\%$ of patterns contain regex metacharacters (\texttt{.}, \texttt{|}, \texttt{()}, \texttt{[]}), so backbones treat \tool{grep} as a regex engine rather than a literal matcher; and the median pattern is $\le 2$ words. The path breakdown confirms the main-text reading: single-file search dominates, with $88$--$99.7\%$ of calls naming one file. The one outlier is Qwen3.5-35B-A3B on BrowseComp, whose $12.0\%$ call-level cross-cache share is driven by two stuck trajectories that repeatedly spam path-less \tool{grep}; measured per question (Table~\ref{tab:cache-tool-calls}), the same backbone issues a whole-cache \tool{grep} on only $7.5\%$ of questions, so the call-level figure overstates how broadly the feature is used.

\begin{table*}[t]
\centering
\small
\caption{\tool{grep} pattern-length distribution (characters) and path-argument breakdown for \FtE{}. \emph{$N$}: total \tool{grep} calls over the benchmark. \emph{x-cache}: share targeting the whole cache (\texttt{path=*}); \emph{single}: share naming one file.}
\label{tab:app-grep}
\setlength{\tabcolsep}{5pt}
\begin{tabular}{ll cccc ccc}
\toprule
& & \multicolumn{4}{c}{Pattern length (chars)} & \multicolumn{3}{c}{Path argument} \\
\cmidrule(lr){3-6} \cmidrule(lr){7-9}
Backbone & Bench & median & p75 & p90 & max & $N$ & x-cache \% & single \% \\
\midrule
\multirow{2}{*}{Qwen3.5-35B-A3B}   & BC & 32 & 46 & 61  & 110 & 590  & 12.0 & 88.0 \\
                                   & WS & 40 & 55 & 93  & 471 & 919  & 2.6 & 97.4 \\
\midrule
\multirow{2}{*}{Qwen3.5-122B-A10B} & BC & 28 & 39 & 53  & 111 & 732  & 2.6 & 97.4 \\
                                   & WS & 29 & 54 & 87  & 534 & 1225 & 4.9 & 95.1 \\
\midrule
\multirow{2}{*}{DeepSeek-V4-Pro}   & BC & 38 & 51 & 66  & 140 & 1133 & 0.3 & 99.7 \\
                                   & WS & 32 & 56 & 93  & 1021 & 2560 & 0.8 & 99.2 \\
\bottomrule
\end{tabular}
\end{table*}

\begin{table*}[t]
\centering
\small
\caption{\tool{read} parameter usage for \FtE{}. \emph{$N$}: total \tool{read} calls. \emph{w/off}, \emph{w/lim}: share of \tool{read} calls carrying an explicit \texttt{offset} / \texttt{limit}. \texttt{offset} (line number) and \texttt{limit} (line count) percentiles are over the calls that specify them.}
\label{tab:app-read}
\setlength{\tabcolsep}{5pt}
\begin{tabular}{ll c cc cc cc}
\toprule
& & & & & \multicolumn{2}{c}{\texttt{offset}} & \multicolumn{2}{c}{\texttt{limit}} \\
\cmidrule(lr){6-7} \cmidrule(lr){8-9}
Backbone & Bench & $N$ & w/off \% & w/lim \% & median & p90 & median & p90 \\
\midrule
\multirow{2}{*}{Qwen3.5-35B-A3B}   & BC & 745  & 39.3 & 56.5 & 101 & 600 & 100 & 200 \\
                                   & WS & 1485 & 44.2 & 80.3 & 200 & 1000 & 150 & 300 \\
\midrule
\multirow{2}{*}{Qwen3.5-122B-A10B} & BC & 825  & 28.5 & 32.2 & 120 & 861 & 100 & 100 \\
                                   & WS & 1269 & 55.7 & 69.7 & 165 & 650  & 100 & 200 \\
\midrule
\multirow{2}{*}{DeepSeek-V4-Pro}   & BC & 1154 & 29.6 & 43.4 & 116 & 810 & 60 & 100 \\
                                   & WS & 1643 & 64.8 & 81.1 & 201 & 607 & 70 & 200 \\
\bottomrule
\end{tabular}
\end{table*}

\paragraph{\tool{read} parameters}
\label{app:supp-read}

Table~\ref{tab:app-read} reports how often \tool{read} carries an explicit \texttt{offset}/\texttt{limit} and their distributions (line-based). Both carry-rates are markedly higher on WideSearch than BrowseComp --- backbones localize by line number more aggressively on the per-field task. The \texttt{offset} distribution shows reads rarely start at line~1 (the strongest backbone's WideSearch p25 is $118$), i.e.\ reads are seeded by \tool{grep}'s returned line numbers. The \texttt{limit} distribution separates strategies: the strongest backbone prefers small, repeated windows (BrowseComp median $60$), the weakest a single large window (WideSearch median $150$, up to $1000$).

\paragraph{Call-sequence $n$-grams}
\label{app:supp-ngram}

Table~\ref{tab:app-ngram} reports the most frequent tool-call bigrams and trigrams (s\,=\,\tool{search}, f\,=\,\tool{fetch}, g\,=\,\tool{grep}, r\,=\,\tool{read}). BrowseComp is dominated by the \tool{search} run s$\to$s$\to$s ($56$--$69\%$ of trigrams); WideSearch surfaces same-type batches (f$\to$f$\to$f, g$\to$g) absent from BrowseComp's top entries. The s$\to$s bigram share falls with backbone strength ($75\% \to 65\%$ on BrowseComp), and the r$\to$s ``read-then-re-search'' edge ($3.3$--$4.6\%$) is the productive feedback loop.

\begin{table*}[t]
\centering
\small
\caption{Most frequent tool-call $n$-grams for \FtE{} (\% of all bigrams / trigrams). s\,=\,\tool{search}, f\,=\,\tool{fetch}, g\,=\,\tool{grep}, r\,=\,\tool{read}.}
\label{tab:app-ngram}
\setlength{\tabcolsep}{4pt}
\begin{tabular}{ll ccc cc}
\toprule
& & \multicolumn{3}{c}{Top bigrams} & \multicolumn{2}{c}{Top trigrams} \\
\cmidrule(lr){3-5} \cmidrule(lr){6-7}
Backbone & Bench & 1st & 2nd & 3rd & 1st & 2nd \\
\midrule
\multirow{2}{*}{Qwen3.5-35B-A3B}   & BC & s$\to$s 75.4 & s$\to$f 6.5  & r$\to$s 3.3 & s$\to$s$\to$s 69.4 & s$\to$s$\to$f 5.6 \\
                                   & WS & s$\to$s 37.3 & f$\to$f 14.3 & s$\to$f 9.0 & s$\to$s$\to$s 30.6 & f$\to$f$\to$f 8.8 \\
\midrule
\multirow{2}{*}{Qwen3.5-122B-A10B} & BC & s$\to$s 72.6 & s$\to$f 7.5  & r$\to$s 3.6 & s$\to$s$\to$s 66.5 & s$\to$s$\to$f 6.0 \\
                                   & WS & s$\to$s 34.5 & f$\to$f 13.4 & s$\to$f 9.9 & s$\to$s$\to$s 28.4 & f$\to$f$\to$f 8.1 \\
\midrule
\multirow{2}{*}{DeepSeek-V4-Pro}   & BC & s$\to$s 64.5 & s$\to$f 9.6  & f$\to$g 4.5 & s$\to$s$\to$s 56.0 & s$\to$s$\to$f 8.3 \\
                                   & WS & s$\to$s 40.4 & f$\to$f 10.8 & g$\to$g 9.7 & s$\to$s$\to$s 33.3 & s$\to$s$\to$f 6.3 \\
\bottomrule
\end{tabular}
\end{table*}

\begin{table*}[t]
\centering
\small
\caption{\tool{fetch} URL reuse for \FtE{}. \emph{Repeat \%}: share of fetches hitting an already-cached URL. \emph{Unique/q}, \emph{Repeats/q}: per-question unique and repeated fetches.}
\label{tab:app-reuse}
\setlength{\tabcolsep}{5pt}
\begin{tabular}{ll ccccc}
\toprule
Backbone & Bench & Fetches & Unique & Repeat \% & Unique/q & Repeats/q \\
\midrule
\multirow{2}{*}{Qwen3.5-35B-A3B}   & BC & 1084 & 1042 & 3.9 & 5.2  & 0.21 \\
                                   & WS & 2244 & 2180 & 2.9 & 10.9 & 0.32 \\
\midrule
\multirow{2}{*}{Qwen3.5-122B-A10B} & BC & 1461 & 1394 & 4.6 & 7.0  & 0.34 \\
                                   & WS & 2257 & 2218 & 1.7 & 11.1 & 0.20 \\
\midrule
\multirow{2}{*}{DeepSeek-V4-Pro}   & BC & 2149 & 2129 & 0.9 & 10.6 & 0.10 \\
                                   & WS & 3053 & 3011 & 1.4 & 15.1 & 0.21 \\
\bottomrule
\end{tabular}
\end{table*}

\paragraph{Fetch reuse}
\label{app:supp-reuse}

Table~\ref{tab:app-reuse} reports how often a \tool{fetch} re-fetches an already-cached URL. Repeat rates are low everywhere (under $5\%$) and lowest for the strongest backbone ($\le 1.4\%$), so the near-zero \tool{list\_fetched} usage, between $0.02$ and $0.15$ calls per question across the six cells, costs little: backbones avoid redundant fetches from context memory without enumerating the cache.

\end{document}